\documentclass{article}

\usepackage{microtype}
\usepackage{graphicx}
\usepackage{subcaption}
\usepackage{booktabs}
\usepackage{hyperref}
\usepackage{tikz}
\usetikzlibrary{positioning,arrows.meta,fit,calc}
\usepackage{multirow}
\usepackage{titletoc}
\usepackage{etoc}
\usepackage{minitoc}

\usepackage[accepted]{icml2026}

\usepackage{amsmath}
\usepackage{amssymb}
\usepackage{mathtools}
\usepackage{amsthm}

\usepackage[capitalize,noabbrev]{cleveref}

\theoremstyle{plain}
\usepackage{makecell}

\theoremstyle{definition}

\theoremstyle{remark}

\icmltitlerunning{Mechanistic Evidence for Preserved-but-Misaligned Representations in Non-IID FedAvg}

\begin{document}

\twocolumn[
\icmltitle{Mechanistic Evidence for Preserved-but-Misaligned Representations in Non-IID FedAvg}

\begin{icmlauthorlist}
\icmlauthor{Muhammad Haseeb}{lums}
\icmlauthor{Salaar Masood}{lums}
\icmlauthor{Muhammad Abdullah Sohail}{lums}
\icmlauthor{Mohammad Fatim Shoaib}{lums}
\icmlauthor{Muhammad Tahir}{lums}
\end{icmlauthorlist}
\icmlaffiliation{lums}{Department of Computer Science, Lahore University of Management Sciences, Lahore, Pakistan}
\icmlcorrespondingauthor{Muhammad Haseeb}{26100253@lums.edu.pk}
\icmlkeywords{Federated Learning, Mechanistic Interpretability, Circuit Analysis, Sparse Autoencoders, Linear Probing, Non-IID Data, Federated Averaging}

\vskip 0.3in
]

\printAffiliationsAndNotice{}

\begin{abstract}
Federated Averaging (FedAvg) often degrades under non-IID client data, but it remains unclear whether this degradation reflects the loss of client-learned representations or a failure to use representations that are still present. We study this question mechanistically in sparse client-trained vision models, using dense-model controls to test whether the observed effects depend on sparsity. Our analysis combines class-specific circuit discovery, linear probing of frozen representations, head-only finetuning, and sparse feature dictionaries. Across CNN and ResNet models on CIFAR-10 and Fashion-MNIST, severe label skew can drive some per-class accuracies near zero even when class-specific internal structure remains recoverable. Linear probes substantially outperform the aggregated classifier, head-only finetuning partially restores accuracy, and USAE transfer reveals a largely shared feature basis between IID and non-IID models. Together, these diagnostics suggest that, in our setting, non-IID FedAvg degradation is not fully explained by representational erasure; it also reflects misalignment between preserved internal structure and the final prediction pathway. We release code, trained checkpoints, extracted circuits, and experiment logs at the following repository:
\href{https://github.com/ha405/FedMI/tree/icml}{FedMI}.
\end{abstract}

\section{Introduction}
\label{sec:introduction}

Federated Learning (FL) trains a shared model from decentralized client data, typically using Federated Averaging (FedAvg)~\citep{mcmahan2017communication,kairouz2021advances}. At each round, clients train local copies of the current global model and the server averages their returned parameters. FedAvg is simple and communication-efficient, but it degrades sharply under non-IID client data, especially label skew~\citep{zhao2018federated,hsieh2020non}. Existing explanations emphasize client drift, objective inconsistency, and weight divergence~\citep{zhao2018federated,karimireddy2020scaffold,wang2020tackling}; these characterize the optimization failure, but leave the internal effect of aggregation largely unobserved. Thus, even when FedAvg succeeds or fails in aggregate, it remains difficult to tell whether client-learned structure was erased, partially preserved, or preserved but poorly used by the final prediction pathway.

\begin{figure}[t]
\centering
\resizebox{0.96\columnwidth}{!}{
\begin{tikzpicture}[
    font=\scriptsize,
    server/.style={
        draw=black!58,
        fill=gray!5,
        rounded corners=2pt,
        align=center,
        minimum width=3.4cm,
        minimum height=2.1cm,
        inner sep=3pt
    },
    client/.style={
        draw=black!42,
        fill=gray!2,
        rounded corners=2pt,
        align=center,
        minimum width=1.50cm,
        minimum height=1.32cm,
        inner sep=3pt
    },
    data/.style={
        draw=black!30,
        fill=gray!4,
        rounded corners=2pt,
        align=center,
        minimum width=1.02cm,
        minimum height=0.30cm,
        inner sep=2pt,
        font=\scriptsize
    },
    down/.style={-{Latex[length=1.15mm]}, line width=0.38pt, draw=black!30},
    up/.style={-{Latex[length=1.15mm]}, line width=0.48pt, draw=black!58},
    edge/.style={draw=black!20, line width=0.27pt},
    unit/.style={circle, draw=black!55, fill=white, minimum size=3.7pt, inner sep=0pt},
    blueunit/.style={circle, draw=blue!55!black, fill=blue!18, minimum size=4.4pt, inner sep=0pt},
    greenunit/.style={circle, draw=green!45!black, fill=green!20, minimum size=4.4pt, inner sep=0pt},
    redunit/.style={circle, draw=red!55!black, fill=red!18, minimum size=4.4pt, inner sep=0pt}
]

\node[font=\tiny, text=black!62] at (0,3.2) {FedAvg Aggregation};
\node[server] (server) at (0,1.95) {};
\node[font=\scriptsize] at (0,2.8) {Server global model};

\draw[dashed, draw=black!30, line width=0.6pt] (0.40, 1.2) -- (0.40, 2.5);
\node[font=\tiny, text=black!60, align=center] at (-0.5, 1.15) {Preserved Circuits};
\node[font=\tiny, text=black!60, align=center] at (0.8, 1.15) {Head};

\foreach \ya in {2.2, 1.85, 1.5} {
  \foreach \yb in {2.2, 1.85, 1.5} {
    \draw[edge] (-1.0,\ya) -- (-0.5,\yb);
    \draw[edge] (-0.5,\ya) -- (0.0,\yb);
  }
}

\foreach \ya in {2.2, 1.85, 1.5} {
    \foreach \yb in {2.2, 1.85, 1.5} {
        \draw[edge, draw=black!10] (0.0, \ya) -- (0.8, \yb);
    }
}

\draw[blue!50, line width=0.8pt] (-1.0, 2.2) -- (-0.5, 1.5) -- (0.0, 1.85);
\draw[green!50!black, line width=0.8pt] (-1.0, 1.85) -- (-0.5, 1.85) -- (0.0, 2.2);
\draw[red!50, line width=0.8pt] (-1.0, 1.5) -- (-0.5, 2.2) -- (0.0, 1.5);

\draw[draw=black!70, line width=0.8pt, -{Latex[length=1.5mm]}] (0.0, 1.85) -- (0.8, 1.5);
\draw[draw=black!70, line width=0.8pt, -{Latex[length=1.5mm]}] (0.0, 2.2) -- (0.8, 2.2);
\draw[draw=black!70, line width=0.8pt, -{Latex[length=1.5mm]}] (0.0, 1.5) -- (0.8, 1.85);

\foreach \x in {-1.0, -0.5, 0.0} {
  \foreach \y in {2.2, 1.85, 1.5} {
    \node[unit] at (\x,\y) {};
  }
}

\node[blueunit] at (-1.0, 2.2) {};
\node[blueunit] at (-0.5, 1.5) {};
\node[blueunit] at (0.0, 1.85) {};

\node[greenunit] at (-1.0, 1.85) {};
\node[greenunit] at (-0.5, 1.85) {};
\node[greenunit] at (0.0, 2.2) {};

\node[redunit] at (-1.0, 1.5) {};
\node[redunit] at (-0.5, 2.2) {};
\node[redunit] at (0.0, 1.5) {};

\node[unit, draw=blue!55!black, thick, fill=white]  at (0.8, 2.2) {};
\node[unit, draw=green!45!black, thick, fill=white] at (0.8, 1.85) {};
\node[unit, draw=red!55!black, thick, fill=white]   at (0.8, 1.5) {};

\node[client] (c1) at (-2.30,-0.15) {};
\node[client] (c2) at (0,-0.15) {};
\node[client] (ck) at (2.65,-0.15) {};

\node[font=\scriptsize] at (-2.30,0.38) {Client 1};
\node[font=\scriptsize] at (0,0.38) {Client 2};
\node[font=\scriptsize] at (2.65,0.38) {Client $K$};

\node[data] at (-2.30,-1.02) {$D_1$};
\node[data] at (0,-1.02) {$D_2$};
\node[data] at (2.65,-1.02) {$D_K$};

\node[font=\large, text=black!55] at (1.32,-0.12) {$\cdots$};

\draw[down] (server.south west) -- (c1.north);
\draw[down] (server.south) -- (c2.north);
\draw[down] (server.south east) -- (ck.north);

\draw[up] (c1.north east) -- (server.south west);
\draw[up] (c2.north) -- (server.south);
\draw[up] (ck.north west) -- (server.south east);

\foreach \ya in {0.08,-0.13,-0.34} {
  \foreach \yb in {0.08,-0.13,-0.34} {
    \draw[edge] (-2.60,\ya) -- (-2.30,\yb);
    \draw[edge] (-2.30,\ya) -- (-2.00,\yb);
  }
}
\draw[blue!50, line width=0.8pt] (-2.60, 0.08) -- (-2.30, -0.34) -- (-2.00, -0.13);
\foreach \x in {-2.60,-2.30,-2.00} {
  \foreach \y in {0.08,-0.13,-0.34} { \node[unit] at (\x,\y) {}; }
}
\node[blueunit] at (-2.60, 0.08) {};
\node[blueunit] at (-2.30, -0.34) {};
\node[blueunit] at (-2.00, -0.13) {};

\foreach \ya in {0.08,-0.13,-0.34} {
  \foreach \yb in {0.08,-0.13,-0.34} {
    \draw[edge] (-0.30,\ya) -- (0,\yb);
    \draw[edge] (0,\ya) -- (0.30,\yb);
  }
}
\draw[green!50!black, line width=0.8pt] (-0.30, -0.13) -- (0, -0.13) -- (0.30, 0.08);
\foreach \x in {-0.30,0,0.30} {
  \foreach \y in {0.08,-0.13,-0.34} { \node[unit] at (\x,\y) {}; }
}
\node[greenunit] at (-0.30, -0.13) {};
\node[greenunit] at (0, -0.13) {};
\node[greenunit] at (0.30, 0.08) {};

\foreach \ya in {0.08,-0.13,-0.34} {
  \foreach \yb in {0.08,-0.13,-0.34} {
    \draw[edge] (2.35,\ya) -- (2.65,\yb);
    \draw[edge] (2.65,\ya) -- (2.95,\yb);
  }
}
\draw[red!50, line width=0.8pt] (2.35, -0.34) -- (2.65, 0.08) -- (2.95, -0.34);
\foreach \x in {2.35,2.65,2.95} {
  \foreach \y in {0.08,-0.13,-0.34} { \node[unit] at (\x,\y) {}; }
}
\node[redunit] at (2.35, -0.34) {};
\node[redunit] at (2.65, 0.08) {};
\node[redunit] at (2.95, -0.34) {};

\node[font=\tiny, text=black!62] at (0,-1.45)
{non-IID clients can learn different internal pathways};

\end{tikzpicture}
}
\caption{FedAvg under non-IID data. Clients train on local datasets $D_k$ and return updates to the server; highlighted units schematically indicate client-specific internal pathways. In the aggregated global model, these functional features are successfully preserved but become misaligned with the final classification head.}
\label{fig:fedavg_intro}
\end{figure}
In this work, we use Mechanistic Interpretability (MI) tools to investigate how FedAvg behaves under non-IID client data. MI studies neural networks by localizing the internal components and representations that implement model behavior~\citep{olah2020zoom,elhage2021mathematical}. We use this perspective to examine how aggregation changes the class-relevant structure learned by local clients, and whether that structure remains recoverable in the global model after FedAvg.

We analyze models trained under IID and
Dirichlet non-IID partitions~\citep{zhao2018federated,hsieh2020non} using three mechanistic
diagnostics. First, we use circuit discovery~\citep{gao2025weightsparse,conmy2023towards} to measure how class-specific subnetworks diverge across clients and how much of their structure is preserved after aggregation. Second, we use linear probes~\citep{alain2016understanding} and head-only finetuning~\citep{oh2022fedbabu} to test whether class information remains accessible in frozen global representations. Third, we apply Universal Sparse Autoencoders (USAEs), building on sparse dictionary learning~\citep{elhage2022toy,bricken2023towards,cunningham2023sparse}, to test whether the IID and non-IID global models share the same underlying feature basis.

Empirically, we observe that non-IID heterogeneity
affects both circuit structure and readout alignment.
Client circuits become less similar as heterogeneity
increases, and the global model only partially
preserves local circuit topology. At the same time,
this same global model retains recoverable class-relevant
structure: extracted circuits remain highly sufficient
even for classes with near-zero global accuracy,
linear probes and head-only finetuning recover
substantial performance under severe skew. Further,
USAE analysis indicates that IID and non-IID models share
most of their feature basis. These results suggest
that non-IID FedAvg degradation involves not only
weakened internal structure, but also misalignment
between retained representations and the model's
default prediction pathway.

\section{Related Work}
\label{sec:related_work}

\subsection{Federated Optimization Under Non-IID Data}

FedAvg remains the canonical optimization procedure for FL, but its local-update-then-average structure is sensitive to statistical heterogeneity~\citep{mcmahan2017communication,kairouz2021advances}. Under non-IID client data, local objectives can drift away from the global objective, producing unstable convergence and degraded global accuracy~\citep{zhao2018federated,hsieh2020non}. A large body of work mitigates this problem by modifying the local objective or the aggregation rule. FedProx adds a proximal term to limit local deviation from the global model~\citep{li2020federated}; SCAFFOLD uses control variates to correct client drift~\citep{karimireddy2020scaffold}; FedNova normalizes local updates to reduce objective inconsistency~\citep{wang2020tackling}; and MOON uses contrastive representation regularization to stabilize local training~\citep{li2021model}. These methods improve optimization behavior, but generally evaluate success through aggregate loss or accuracy. Our work instead asks how non-IID aggregation changes the internal structure of the learned model.

\subsection{Classifier and Readout Mismatch in Federated Learning}

Several FL methods suggest that the classifier head is especially vulnerable under heterogeneous label distributions. FedRS studies label-distribution skew and argues that the last classification layer is more sensitive to missing or imbalanced classes than lower feature layers~\citep{li2021fedrs}. Personalized FL methods make a related architectural separation: FedPer keeps shared lower layers while personalizing upper layers~\citep{arivazhagan2019federated}, FedRep learns a shared representation with client-specific heads~\citep{collins2021exploiting}, and FedBABU updates the model body during federated training while adapting the head afterward~\citep{oh2022fedbabu}. These methods improve optimization behavior, but leave open how non-IID aggregation changes the internal structure of the learned model.

\subsection{Model Merging and Weight-Space Averaging}

FedAvg can also be viewed as a repeated form of weight-space model merging, where locally trained models with a shared initialization are averaged into a single global model. Recent model-merging work shows that parameter averaging and task-vector arithmetic can combine capabilities when models remain sufficiently aligned in weight space or lie in compatible loss basins~\citep{wortsman2022model,matena2022merging,ilharco2023editing}. At the same time, work on mode connectivity, model re-basing, and interference-aware merging emphasizes that permutation symmetries, representation alignment, and conflicting parameter directions can strongly affect whether independently trained models can be merged successfully~\citep{frankle2020linear,ainsworth2023git,yadav2023ties}. These results provide a useful lens on FedAvg: non-IID clients may learn partially incompatible internal structures before aggregation. We use this perspective to study FedAvg at the level of circuits, probes, and sparse feature dictionaries rather than only post-merge accuracy.

\subsection{Mechanistic Interpretability, Circuits, and Sparse Features}

Mechanistic Interpretability aims to explain model behavior by identifying the internal components and computations that implement it~\citep{olah2020zoom,elhage2021mathematical}. Circuit analysis studies sparse subsets of model components sufficient for particular behaviors~\citep{olah2020zoom,wang2022interpretability}, with work on transformer circuits, automated circuit discovery, and circuit faithfulness establishing increasingly systematic tools for localizing mechanisms~\citep{elhage2021mathematical,conmy2023towards,hanna2024faithfulness}. Sparse models can further make circuit extraction more tractable; recent work shows that weight-sparse transformers can expose interpretable task circuits~\citep{gao2025weightsparse}. In parallel, Sparse Autoencoders (SAEs) address superposition by decomposing dense activations into sparse feature directions~\citep{elhage2022toy,bricken2023towards,cunningham2023sparse}. We bring these tools into the federated setting, using circuits to study structural preservation, probes to test activation-level accessibility, and USAEs to compare IID and non-IID feature spaces, with the goal of distinguishing representational erasure from preserved-but-misaligned structure after FedAvg.

\section{Methodology}
\label{sec:methods}

\begin{figure*}[t]
  \centering
  \includegraphics[width=0.98\textwidth]{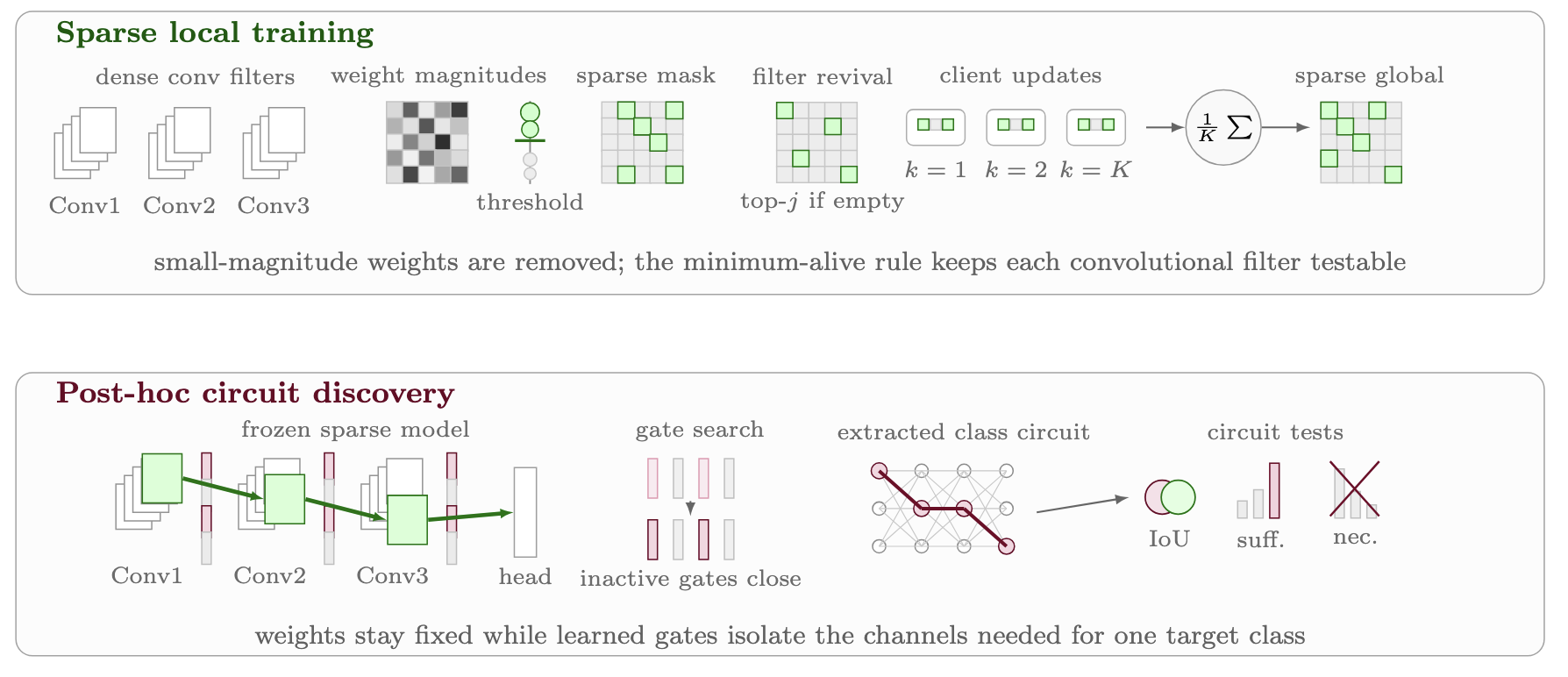}
  \caption{Sparse training and circuit discovery. During sparse local training, magnitude pruning removes small weights while a minimum-alive constraint revives the top-$j$ weights of any dead filter; sparse client models are then aggregated by FedAvg. Circuit discovery is performed after training by freezing the model and learning channel gates that isolate a class-specific circuit, which is evaluated by IoU, sufficiency, and necessity.}
  \label{fig:method_overview}
\end{figure*}

\subsection{Federated Learning Setup}

We evaluate FedAvg~\citep{mcmahan2017communication} on two datasets, CIFAR-10 and Fashion-MNIST, and two architectures, SimpleCNN and ResNet10. SimpleCNN consists of three convolutional blocks with channel widths $[64,128,256]$, max-pooling after each block, and a final linear classification head. ResNet10 consists of four sequential BasicBlocks with widths $[64,128,192,256]$, skip connections within each block, global average pooling, and a final linear head. Detailed layer specifications, output sizes, and circuit-gated modules for both architectures are reported in Appendix~\ref{app:model_arch_extra}.

Unless otherwise noted, all experiments use $N=10$ clients, 5 local epochs per round, batch size 256, Adam with learning rate $10^{-3}$, and 100 communication rounds. At each round, clients initialize from the current global model, perform local training, and the server aggregates the resulting parameters by uniform averaging.

To vary statistical heterogeneity, we partition the training data across clients using a Dirichlet distribution with concentration parameter $\alpha \in \{0.5, 0.2, 0.05\}$, alongside an IID baseline.


\subsection{Sparse Local Training}
\label{subsec:sparse_training}

Client models are trained under a global magnitude-based sparsity schedule that anneals the target sparsity from $0$ to $\rho=99.9\%$ during local training~\citep{gao2025weightsparse}. Let $T_{\mathrm{local}}$ denote the number of local optimization steps in a client update and let $T_{\mathrm{ramp}}=\lceil 0.5T_{\mathrm{local}}\rceil$. At local step $t$, we set
\begin{equation}
\label{eq:sparsity_schedule}
    \rho_t = \rho \cdot \min\!\left(\frac{t}{T_{\mathrm{ramp}}}, 1\right),
\end{equation}
so sparsity increases linearly over the first half of local training and is then held fixed at $\rho$. The magnitude threshold $\tau_t$ is chosen so that exactly $\lceil (1-\rho_t)|\theta| \rceil$ scalar weights are retained.

Thus, our main FedAvg experiments operate in a sparse-training regime. We use this setting deliberately because weight-sparse models yield compact, interventionally testable circuits, and prior bridge experiments suggest that sparse models can expose mechanisms relevant to dense models~\citep{gao2025weightsparse}. Importantly, we also run dense-model controls in Appendix~\ref{app:dense_model_justification}, where local pruning is disabled and the same post-hoc analyses are repeated. The same qualitative circuit divergence, representational recovery, and feature-sharing patterns persist there, so sparsity is used as an interpretability-enabling lens rather than as a necessary condition for the paper's central finding.

To avoid fully pruned convolutional filters, we impose a minimum-alive constraint. If an output filter is left with fewer than $j=4$ active weights after magnitude pruning, we replace the corresponding mask by its top-$j$ weights in magnitude~\citep{gao2025weightsparse}:

\begin{equation}
\label{eq:revival}
m'_o =
\begin{cases}
    \mathbf{1}\{|w_o| \ge \operatorname{top}_j(|w_o|)\} & \text{if } \|m_o\|_0 < j \\
    m_o & \text{otherwise}
\end{cases}
\end{equation}

where $\mathbf{1}\{\cdot\}$ denotes an elementwise binary indicator mask, $\|m_o\|_0$ denotes the number of active weights in the original mask, and $\operatorname{top}_j(|w_o|)$ denotes the magnitude of the $j$th largest weight in filter $o$.

\subsection{Circuit Discovery}
\label{subsec:circuit_discovery}

Following the circuit-finding methodology of~\citet{gao2025weightsparse}, we
define the \textbf{circuit} for class $c$ as the minimal sub-network of
parameters that is sufficient to correctly predict class $c$. Circuits are
discovered via differentiable binary gating: for each eligible convolutional and
linear layer, we attach per-output-channel gate scalars
$g_{l,j} \in \mathbb{R}$, where $l$ indexes the gated layer and $j$ indexes an output channel or linear unit within that layer. Gates are initialised to $2.0$. During the forward pass, layer
$l$'s output is multiplied channel-wise by a straight-through binary estimator:
\begin{equation}
    \hat{g}_{l,j} \;=\;
    \underbrace{\mathbf{1}\{g_{l,j}>0\}}_{\text{binary, forward}}
    \;-\; \underbrace{\sigma(g_{l,j}) - \sigma(g_{l,j})}_{\text{zero}}
    \;+\; \underbrace{\sigma(g_{l,j})}_{\text{gradient path}},
\end{equation}
so that the backward pass receives gradients through $\sigma(g_{l,j})$ while the
forward pass uses a hard binary gate~\citep{bengio2013estimating}. Gate
parameters are optimised jointly over the cross-entropy loss and an $\ell_0$
sparsity penalty:
\begin{equation}
    \mathcal{L} \;=\;
    \mathcal{L}_{\mathrm{CE}}
    \;+\;
    \lambda_0 \sum_{l}\sum_{j} \sigma(g_{l,j}),
\end{equation}
with $\lambda_0{=}0.01$ and $T{=}300$ optimisation steps using Adam at learning
rate $0.1$. After optimisation, gate $g_{l,j}$ is deemed active if
$g_{l,j}>0$, and the set of active output-channel indices per layer constitutes
the circuit for class $c$.

To ensure comparability across clients, circuits are extracted using a
\textbf{balanced global discovery pool} drawn uniformly from the full training
dataset (20\% per class, up to 1{,}024 samples per class).

\paragraph{Circuit Interventions.}
After a class circuit is extracted, we evaluate its functional role with two
interventions on the frozen model. \textbf{Sufficiency} keeps only the active
gated channels in the circuit and masks the remaining eligible channels; the
reported score is the target-class accuracy of this circuit-only model.
\textbf{Necessity} performs the complementary intervention: active circuit
channels are ablated from the otherwise unmodified model, and the reported score
is the remaining target-class accuracy. A high sufficiency score together with a
low necessity score indicates that the extracted circuit is capable of
supporting the class prediction and that removing it substantially disrupts that
prediction.

\paragraph{Structural Similarity.}
Two circuits $C_a$ and $C_b$ are compared via the mean per-layer
\textbf{Intersection-over-Union}:
\begin{equation}
    \mathrm{IoU}(C_a, C_b)
    \;=\;
    \frac{1}{|L|}\sum_{l \in L}
    \frac{|C_a^l \cap C_b^l|}{|C_a^l \cup C_b^l|},
\end{equation}
where $L$ is the set of eligible layers. We summarize circuit structure with
mean IoU scores, where ``mean'' denotes averaging this per-layer IoU over the
relevant classes, clients, or rounds for a given comparison. We use
two complementary IoU measurements. \textbf{Inter-client
consistency} is the mean IoU across all client pairs and all classes, measuring
whether different clients learn similar class circuits. \textbf{Local-to-global
preservation} is the mean IoU between each client's local circuit and the
corresponding circuit extracted from the aggregated global model after each
communication round, measuring how much client circuit structure survives
FedAvg.

\subsection{Linear Probing}

Linear probing separates representation quality from the global model's default
classifier: if a probe succeeds while the original head fails, then class
information is still present in the frozen features but is not being used
effectively by the learned readout. We train a linear probe on frozen
penultimate-layer activations of the global model using the full training set,
for 50 epochs with Adam at learning rate $10^{-4}$ and batch size 128. We compare
probe accuracy against the global model's own classification accuracy; the
resulting gap serves as a diagnostic for representational accessibility versus
classifier misalignment.

\subsection{Classification Head Finetuning}

To test whether part of the accuracy degradation is localized to the final readout, we freeze the backbone of the global model and finetune only the final linear classification head on a balanced IID subset of 200 training samples (20 per class). The head is optimized for 5 epochs using AdamW with learning rate $10^{-3}$, weight decay $5 \times 10^{-4}$, batch size 16, and cosine annealing. We compare post-finetuning accuracy against the original global model across heterogeneity levels; the IID model serves as a control.


\subsection{Universal Sparse Autoencoder}

Probes and head finetuning test label accessibility, but they do not directly
show whether IID and non-IID models organize their features in a shared basis.
To compare feature structure between IID and non-IID global models, we train a
USAE~\citep{bricken2023towards,cunningham2023sparse} jointly on the
penultimate-layer activations of both models. The USAE uses model-specific
encoders and decoders together with a shared dictionary of $K=2048$ concepts
($\sim 8\times$ expansion). Activations are extracted from both models on the
same held-out test loader and stored as paired activation batches
$(A_{\mathrm{IID}}, A_{\mathrm{nonIID}})$. During USAE training, each minibatch
randomly chooses either model as the source side, so the dictionary is learned
from both IID and non-IID activations rather than from one model alone. We treat
a concept as shared if it activates non-trivially for both models over this
paired activation pool. To assess cross-model compatibility, we evaluate concept
transfer by encoding activations with the source encoder, decoding them into the
target activation space, and classifying them with the target head.

\section{Results}
\label{sec:results}
We evaluate global models trained under IID and 
Non-IID client partitions on CIFAR-10 with the CNN 
backbone, with concentration parameter 
$\alpha \in \{0.5, 0.2, 0.05\}$. Our analysis 
follows a diagnostic chain: we first measure how 
Non-IID training affects circuit structure within 
clients, across clients, and after aggregation. We 
then test whether class-relevant structure remains 
recoverable despite degraded global accuracy, using 
circuit extraction, linear probing, head-only 
finetuning, and feature subspace analysis. 
Appendix~\ref{app:circuit_extra}--\ref{app:usae_extra} 
report the same analyses for CIFAR-10 with ResNet 
and Fashion-MNIST with both backbones.

\subsection{Circuit Stability and Divergence}
\label{subsec:circuit_stability}
We examine circuits from two complementary angles: 
whether they stabilize \emph{within} each client 
across rounds (intra-client), and whether clients 
learn similar circuits \emph{across} each other 
(inter-client).

\begin{figure}[ht]
  \vskip 0.2in
  \begin{center}
    \centerline{\includegraphics[width=\columnwidth]
    {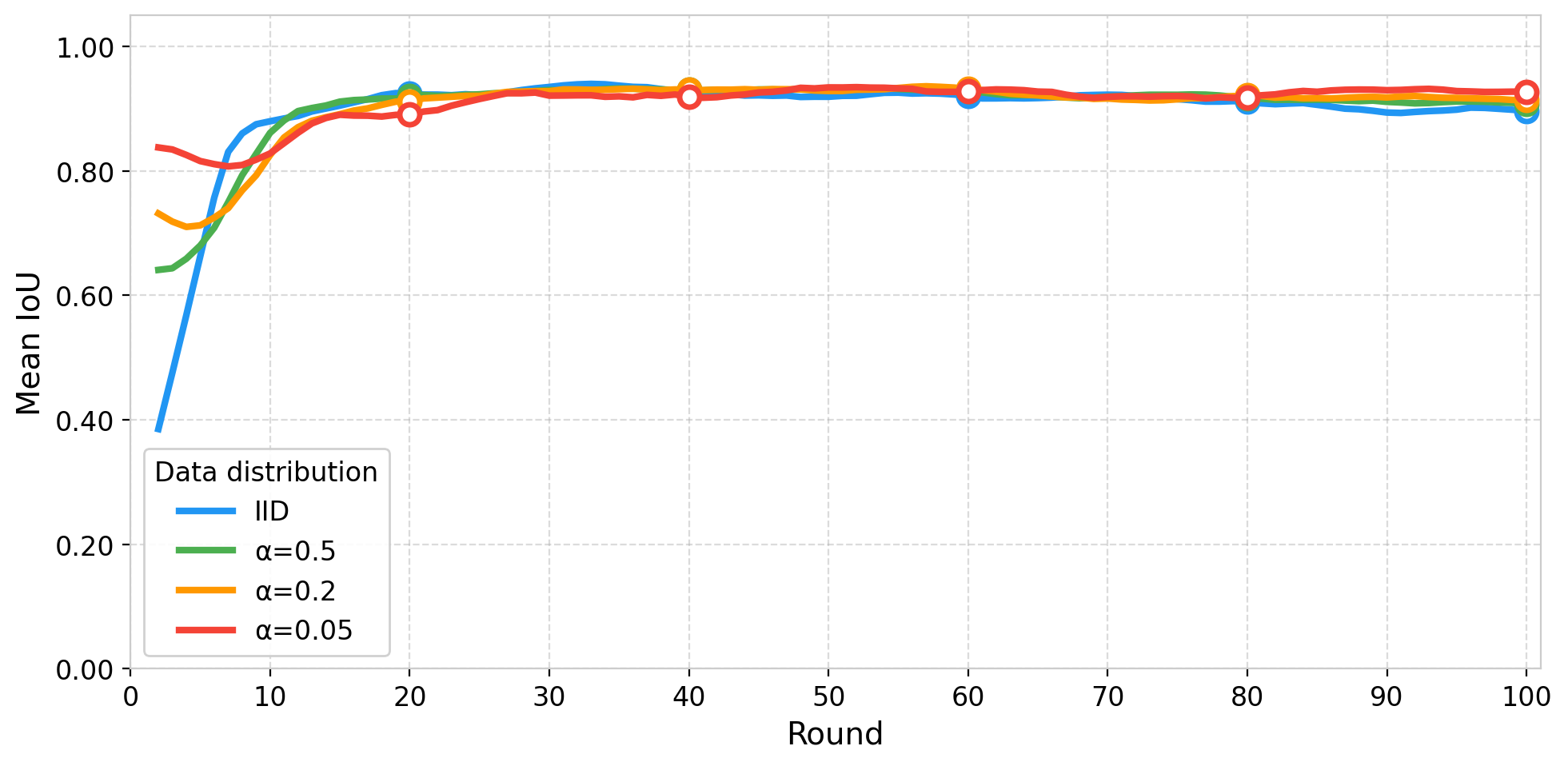}}
    \caption{Intra-client circuit stability across communication rounds. Consecutive-round circuit overlap remains high across all client data distributions, stabilizing around IoU $\approx 0.85$--$0.92$ for both IID and highly skewed non-IID partitions.}

    \label{fig:intra_circuit_stability}
  \end{center}
  \vskip -0.2in
\end{figure}

As shown in Figure~\ref{fig:intra_circuit_stability}, local training dynamics remain stable regardless of the client data distribution. Even under extreme non-IID conditions ($\alpha = 0.05$), the structural overlap between consecutive rounds rapidly climbs and plateaus near 0.90. This demonstrates that the failure of non-IID FedAvg is not rooted in local training instability or noise. Instead, local clients successfully converge and lock into highly stable computational pathways for the specific data they observe.

\begin{figure}[ht]
  \vskip 0.2in
  \begin{center}
    \centerline{\includegraphics[width=\columnwidth]
    {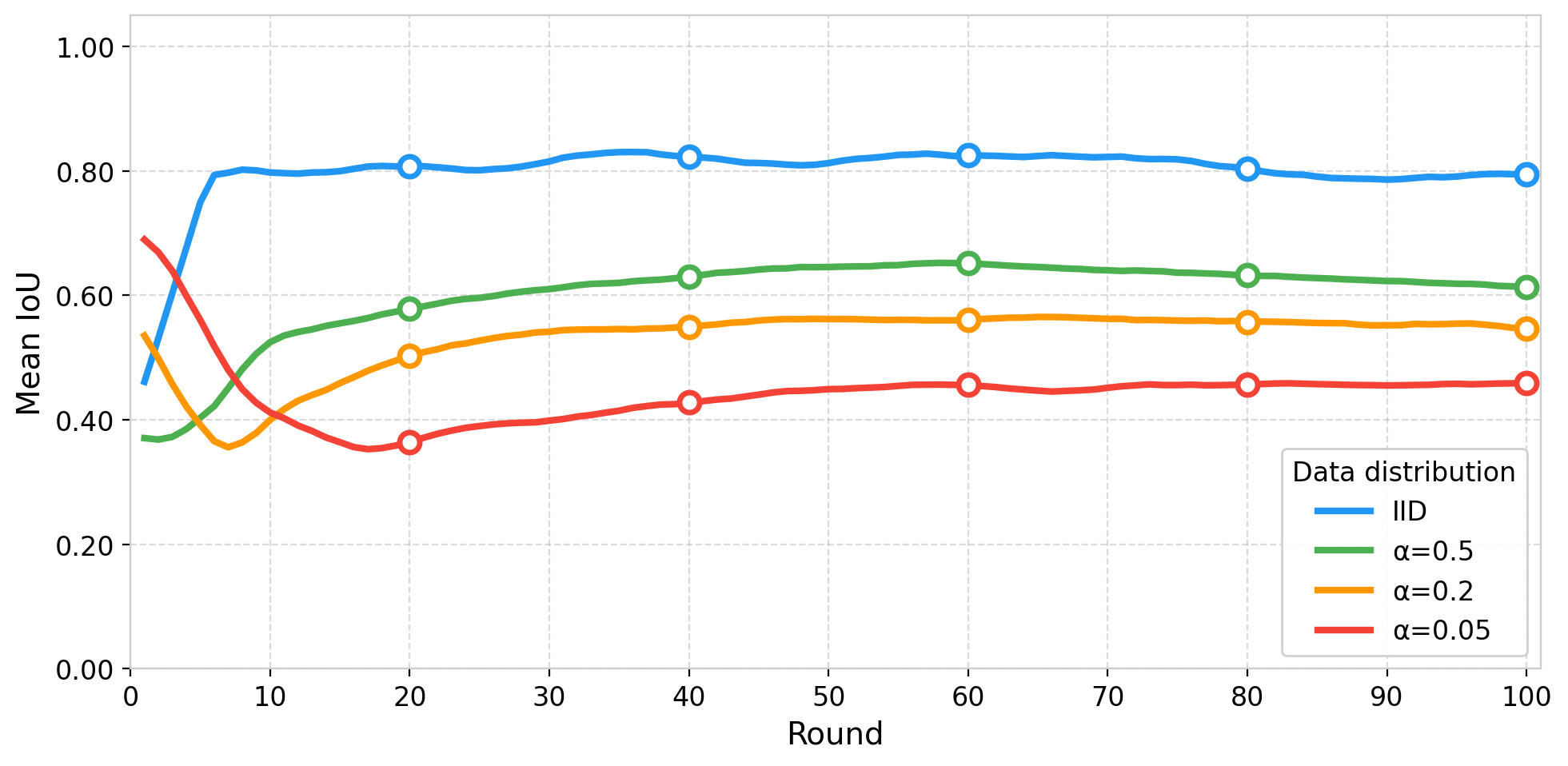}}
    \caption{Inter-client circuit consistency across communication rounds. Under IID partitions, clients converge to similar class-specific circuits (IoU $\approx 0.80$). As heterogeneity increases, client circuits diverge, stabilizing near IoU $\approx 0.45$ at $\alpha = 0.05$.}

    \label{fig:inter_circuit_consistency}
  \end{center}
  \vskip -0.2in
\end{figure}
\textbf{Inter-client divergence.} Under IID 
conditions, all clients converge to similar circuit 
structures and maintain this throughout training. 
As heterogeneity increases, circuits diverge between 
clients. Under extreme skew ($\alpha = 0.05$), 
clients rely on substantially different circuits 
to represent the same class, with structural overlap 
stabilizing near IoU $\approx 0.45$.

Together, these results show that federated training 
produces stable local circuits regardless of data 
heterogeneity, but what clients converge to depends 
heavily on their local class distribution, causing
clients to stabilize around different class-specific
circuits.

\subsection{Local vs Global Circuits}
\label{subsec:local_vs_global}

Given that Non-IID clients learn divergent circuits,
we next ask how FedAvg integrates these conflicting
structures. We measure the IoU between each client's
local circuit and the corresponding circuit extracted
from the aggregated global model using the
local-to-global preservation metric defined in
Section~\ref{subsec:circuit_discovery}.

\begin{figure}[ht]
  \vskip 0.2in
  \begin{center}
    \centerline{\includegraphics[width=\columnwidth]
    {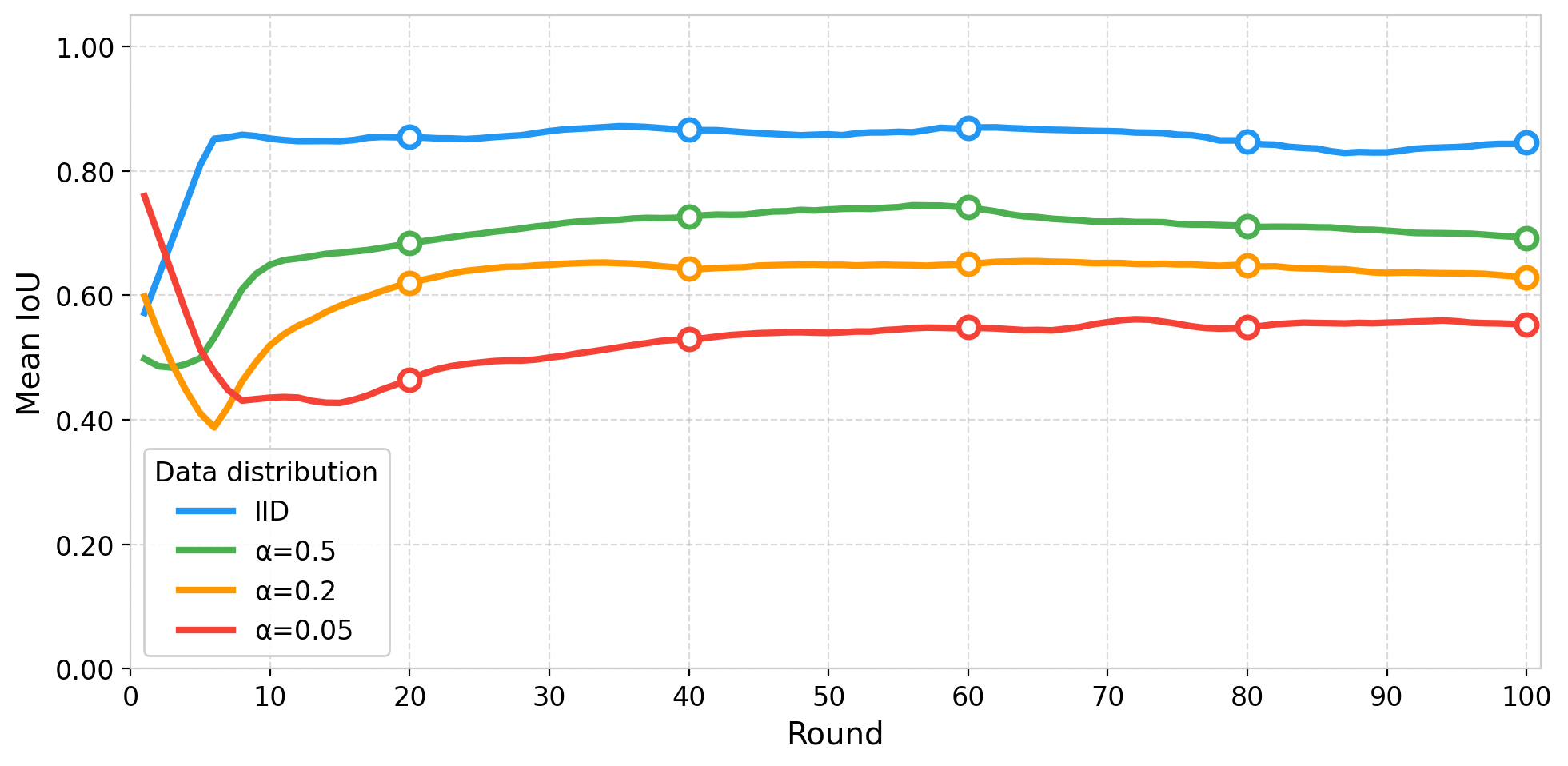}}
    \caption{Local-to-global circuit preservation after FedAvg. The IID global model closely preserves client circuit structure (IoU $\approx 0.85$). Under non-IID partitions, preservation drops but does not collapse: even at $\alpha = 0.05$, the global model retains roughly 0.55 IoU with local client circuits.}

    \label{fig:local_vs_global_consistency}
  \end{center}
  \vskip -0.2in
\end{figure}

Figure~\ref{fig:local_vs_global_consistency} shows that, under IID conditions, the global model preserves local
circuit topology nearly faithfully. Under Non-IID
conditions, preservation degrades but does not
collapse. The key observation is that local-to-global
preservation (IoU $\approx 0.55$) remains higher than
inter-client circuit overlap (IoU $\approx 0.45$) even
under extreme skew. This means FedAvg does not simply
average circuits into noise. The global model inherits partial structure from clients rather than converging to any single local solution. Appendix~\ref{app:circuit_extra} shows the same partial preservation pattern across the remaining dataset and backbone pairs.

\subsection{Functional Circuits Persist Even for
Near-Zero Accuracy Classes}
\label{subsec:circuit_persistence}

Section~\ref{subsec:local_vs_global} shows that the
global circuit only partially overlaps with local
clients' circuits under Non-IID conditions. We now
ask a stronger question: does the global model retain
working circuits for classes it can no longer classify?
Table~\ref{tab:per_class_acc} reports per-class
accuracy, circuit sufficiency, and necessity at the final communication round, round 100, under extreme Non-IID ($\alpha = 0.05$).

\begin{table}[ht]
  \caption{Per-class global accuracy, circuit
  sufficiency, and necessity at round 100 under
  extreme Non-IID ($\alpha = 0.05$) for CIFAR-10
  with CNN. Overall global accuracy is 43.05\%. Nodes denotes
  active gated channels in the extracted global circuit.}
  \label{tab:per_class_acc}
  \centering
  \setlength{\tabcolsep}{5pt} 
  \begin{footnotesize}
  \begin{sc}
  \renewcommand\theadfont{\scshape} 
  \begin{tabular}{lcccc}
    \toprule
    Class & \thead{Global\\Acc} & \thead{Circuit\\Suff.} & Necessity & Nodes \\
    \midrule
    Airplane   &  0.5\% &  99.9\% &  0.0\% & 43 \\
    Automobile & 43.3\% & 100.0\% &  0.0\% & 47 \\
    Bird       & 82.1\% & 100.0\% & 28.0\% & 42 \\
    Cat        &  0.0\% & 100.0\% &  0.0\% & 41 \\
    Deer       &  0.0\% & 100.0\% &  0.0\% & 32 \\
    Dog        & 53.4\% & 100.0\% &  0.1\% & 26 \\
    Frog       & 69.3\% & 100.0\% &  0.0\% & 24 \\
    Horse      & 20.8\% & 100.0\% &  0.0\% & 30 \\
    Ship       & 81.0\% & 100.0\% &  0.0\% & 36 \\
    Truck      & 80.1\% & 100.0\% &  0.0\% & 46 \\
    \bottomrule
  \end{tabular}
  \end{sc}
  \end{footnotesize}
\end{table}

Circuit sufficiency is near-perfect across all ten
classes regardless of global accuracy. Classes at
0\% global accuracy still yield circuits with 100\%
target-class sufficiency. Necessity results confirm
that these circuits are not redundant: ablating the
circuit nodes collapses accuracy to near zero for
nine out of ten classes, with Bird being the only
exception where backup pathways retain 28\% accuracy.
Together, sufficiency and necessity establish that
the discovered circuits are the primary computational
pathway for each class in the global model. The fact
that these circuits exist and are causally responsible
for class prediction, yet the model scores 0\% on
several classes end-to-end, confirms that the failure
is in how the model routes inputs through these
circuits during normal inference, not in whether the
circuits exist. Appendix~\ref{app:circuit_suff_extra}
shows the same pattern across the remaining dataset
and backbone pairs.

\subsection{Linearly Separable Global Representations}\label{subsec:linear_probing}

Circuit discovery searches the weight space for
functional sub-networks. A complementary question is
whether the same latent structure is visible from the
activation side: do the global model's penultimate-layer
representations contain enough class-discriminative
geometry that a simple linear classifier can exploit
it, even when the model's own head cannot? To test
this, we train a single linear layer on the frozen
penultimate-layer activations of the global model
\citep{alain2016understanding} and compare its test
accuracy to the global model's default classification
accuracy.

\begin{table}[ht]
  \caption{Global model accuracy vs.\ linear probe
  accuracy on CIFAR-10. The probe is a single linear
  layer trained on frozen features from the global
  model's penultimate layer for 50 epochs with Adam
  at learning rate $10^{-4}$ and batch size 128.
  Delta is probe accuracy minus global accuracy.}
  \label{tab:linear_probe}
  \centering
  \begin{small}
  \begin{sc}
  \begin{tabular}{lccc}
    \toprule
    Config & Global Acc & Probe Acc & Delta \\
    \midrule
    IID           & 68.3\% & 68.5\% & +0.2\%  \\
    $\alpha=0.5$  & 63.8\% & 67.5\% & +3.7\%  \\
    $\alpha=0.2$  & 55.5\% & 66.9\% & +11.4\% \\
    $\alpha=0.05$ & 43.1\% & 64.7\% & +21.7\% \\
    \bottomrule
  \end{tabular}
  \end{sc}
  \end{small}
\end{table}

Under IID conditions, probe and global accuracy are
nearly identical, as expected. Under Non-IID conditions,
a gap emerges and widens monotonically with
heterogeneity. At $\alpha=0.05$, the gap reaches 21.7
points. Probe accuracy remains stable across all heterogeneity
levels, dropping only 3.8 points from IID to
$\alpha=0.05$, while default global accuracy drops
25.2 points over the same range. The backbone continues to
produce class-discriminative features that a fresh
linear classifier can exploit, even as the global
model's own head fails. Appendix~\ref{app:linear_probe_extra}
shows this pattern holds across all four
architecture and dataset combinations, with probe
gains between 11 and 22 points under extreme skew.

\subsection{Head Realignment}
\label{subsec:head_finetune}

The linear probe shows that class-discriminative
features remain linearly separable in the frozen
backbone. A direct test is whether this is exploitable
in practice: we finetune only the classification head
on 200 balanced samples for 5 epochs while keeping
the backbone frozen. The IID model serves as a
control: its head should already be well-aligned, so
finetuning on 200 samples should yield negligible
change.

\begin{table}[ht]
  \caption{Global model accuracy before and after
  head-only finetuning on 200 balanced IID samples
  for 5 epochs. The backbone is frozen throughout.
  Delta is post-finetune minus pre-finetune accuracy.}
  \label{tab:head_finetune}
  \centering
  \begin{small}
  \begin{sc}
  \begin{tabular}{lccc}
    \toprule
    Config & Pre-FT & Post-FT & Delta \\
    \midrule
    IID           & 68.3\% & 65.9\% & $-2.4\%$ \\
    $\alpha=0.5$  & 63.8\% & 64.0\% & $+0.2\%$ \\
    $\alpha=0.2$  & 55.5\% & 60.9\% & $+5.4\%$ \\
    $\alpha=0.05$ & 43.1\% & 53.3\% & $+10.2\%$ \\
    \bottomrule
  \end{tabular}
  \end{sc}
  \end{small}
\end{table}

The IID model is almost unaffected ($-2.4\%$), confirming
the control. For Non-IID models, recovery scales
monotonically with heterogeneity, reaching $+10.2$
points at $\alpha=0.05$. However, the recovery is partial as it is still 15 points lower than IID baselines. This suggests that backbone geometry also degrades under Non-IID aggregation. Appendix~\ref{app:head_finetune_extra} shows the same pattern across all four model families, with gains ranging from $+8.7$ to $+9.9$ points under severe
heterogeneity.

\subsection{Shared Feature Structure Across IID and
Non-IID Models}
\label{subsec:usae}

The linear probe shows that class-discriminative
features survive aggregation. We now ask whether these
are the \emph{same} features present in the IID model,
or different features that happen to be linearly
decodable. To test this, we train a USAE \citep{bricken2023towards,
cunningham2023sparse} jointly on paired penultimate-layer
activations of the IID global model (68.3\% accuracy)
and the extreme Non-IID global model ($\alpha=0.05$,
43.1\% accuracy), producing a shared concept dictionary
of $K=2048$ entries with model-specific encoders and
decoders.

\textbf{Concept overlap.} Of the 2048 dictionary
entries, 2021 (98.7\%) activate non-trivially on inputs
from both models over the paired held-out activation
pool; this overlap is computed globally over the paired
test-loader activations, not as a per-class average.
The shared and full dictionaries
differ by at most 0.003 accuracy in every condition,
confirming that the 27 model-specific concepts carry
negligible functional information. The two models share
nearly the same feature basis.

\textbf{Cross-model reconstruction.}
Table~\ref{tab:usae_results} reports accuracy under
USAE round-trips: activations encoded by the source
model, decoded into the target model's activation
space, and classified by the target head. Same-model
entries bound the maximum achievable accuracy given
reconstruction loss; cross-model entries test whether
features transfer.

\begin{table}[ht]
  \caption{USAE concept-transfer accuracy on CIFAR-10.
  Same-model entries are reconstruction upper bounds;
  cross-model entries test concept transfer.}
  \label{tab:usae_results}
  \centering
  \begin{small}
  \begin{sc}
  \begin{tabular}{lcc}
    \toprule
    Dictionary & Source $\to$ Target & Accuracy \\
    \midrule
    Shared (2021)
        & IID $\to$ IID         & 0.6421 \\
        & IID $\to$ Non-IID     & 0.3645 \\
        & Non-IID $\to$ IID     & 0.5899 \\
        & Non-IID $\to$ Non-IID & 0.4054 \\
    \midrule
    Full (2048)
        & IID $\to$ IID         & 0.6426 \\
        & IID $\to$ Non-IID     & 0.3650 \\
        & Non-IID $\to$ IID     & 0.5921 \\
        & Non-IID $\to$ Non-IID & 0.4083 \\
    \bottomrule
  \end{tabular}
  \end{sc}
  \end{small}
\end{table}

Two patterns stand out in Table~\ref{tab:usae_results}.

First, shared and full dictionaries perform nearly
identically in every condition. The 27 model-specific
concepts make no meaningful difference.

Second, when Non-IID activations are translated into the IID model’s activation space through the USAE, the IID head still performs well, close to its own
same-model upper bound. The Non-IID head
fails regardless of concept source: it scores 0.4054
on its own activations and drops further to 0.3645
when fed IID activations, both below its unmodified
accuracy of 43.1\%. This is consistent with the
probe and head finetuning results: the features are
present, but the Non-IID classification head cannot
use them. Appendix~\ref{app:usae_extra} shows that
the fraction of shared concepts remains above 91\%
across all four model families.

Further cross-reconstruction analysis using SVCCA confirms that this shared feature basis is preserved in a misaligned coordinate system and we can recover accuracy even when using the original unmodified IID head on Non-IID backbone. We show this in Appendix~\ref{app:svcca_alignment}

\section{Discussion}
\label{sec:discussion}

\textbf{Reframing weight divergence.}
Prior work attributes Non-IID failure to weight
divergence: conflicting updates overwrite learned
representations during averaging
\citep{zhao2018federated, karimireddy2020scaffold}.
Our results do not support this as the primary
mechanism. Across four independent analyses, we find that substantial class-relevant structure remains recoverable from the global backbone after aggregation. What changes is
the structural relationship between the backbone and
the classification head. Existing methods that reduce
update divergence \citep{li2020federated,
karimireddy2020scaffold, wang2020tackling} address a
symptom of this problem. Our findings suggest the
root cause is representational misalignment rather
than representational loss.

\textbf{Implications for algorithm design.}
Two directions follow from these results. First,
post-aggregation head realignment on a small balanced
set directly targets the identified failure mode and
requires no changes to the federated training protocol.
Second, circuit-aware aggregation --- aligning neurons
that implement the same functional circuit across
clients before averaging --- could reduce structural
shift at its source. Neuron permutation methods for
model merging \citep{ainsworth2023git} are applicable
here, with the IoU metric used in this paper providing
a natural objective for measuring alignment quality.
More broadly, mechanistic tools can expose failures
invisible to aggregate accuracy: a model may appear
to have forgotten a class while still containing a
recoverable circuit for it.

\textbf{Limitations.}
Our analysis is conducted on CIFAR-10 and Fashion-MNIST
with CNN and ResNet architectures. Whether these
findings extend to transformer architectures, language
models, or privacy-constrained FL settings remains
open. Circuit discovery operates at the channel level
and does not capture sub-channel structure. The linear
probe confirms linear separability but does not rule
out non-linear representational damage. USAE round-trip
numbers should be read against same-model upper bounds
due to reconstruction loss. The residual 15-point gap
to IID performance after head finetuning at
$\alpha=0.05$ indicates that backbone degradation also
contributes and head realignment alone is not a
complete solution.
\section{Conclusion}
\label{sec:conclusion}

We studied FedAvg under non-IID data through a mechanistic lens, asking whether poor global accuracy reflects destroyed client knowledge or a failure to express knowledge that remains internally present. Across circuit discovery, linear probing, head-only finetuning, and USAE feature analysis, we find consistent evidence for the latter: even when the global model performs poorly on some classes, class-relevant circuits, linearly decodable representations, and shared sparse features remain recoverable from the aggregated model. This pattern appears across dataset--backbone pairs and is further supported by dense-model controls.

These results suggest that non-IID degradation should not be understood only as statistical drift or representational collapse. In our experiments, FedAvg can preserve useful internal structure while disrupting its alignment with the final prediction pathway. This reframes part of the non-IID problem as one of mechanistic preservation and expression, and points toward federated methods that diagnose and repair where recoverable knowledge becomes disconnected from behavior.

\bibliography{references}

@inproceedings{mcmahan2017communication,
  title={Communication-Efficient Learning of Deep Networks from Decentralized Data},
  author={McMahan, H. Brendan and Moore, Eider and Ramage, Daniel and Hampson, Seth and Arcas, Blaise Aguera y},
  booktitle={Proceedings of the 20th International Conference on Artificial Intelligence and Statistics (AISTATS)},
  year={2017},
  volume={54},
  series={Proceedings of Machine Learning Research}
}

@article{zhao2018federated,
  title={Federated Learning with Non-IID Data},
  author={Zhao, Yue and Li, Meng and Lai, Liangzhen and Suda, Naveen and Civin, Damon and Chandra, Vikas},
  journal={arXiv preprint arXiv:1806.00582},
  year={2018}
}

@article{elhage2022toy,
  title={Toy Models of Superposition},
  author={Elhage, Nelson and Hume, Tristan and Olsson, Catherine and Schiefer, Nicholas and Henighan, Tom and Kravec, Shauna and Hatfield-Dodds, Zac and Lasenby, Robert and Drain, Dawn and Chen, Carol and others},
  journal={arXiv preprint arXiv:2209.10652},
  year={2022}
}

@article{gao2025weightsparse,
  title={Weight-sparse transformers have interpretable circuits},
  author={Gao, Leo and Rajaram, Achyuta and Coxon, Jacob and Govande, Soham V. and Baker, Bowen and Mossing, Dan},
  journal={arXiv preprint arXiv:2511.13653},
  year={2025}
}

@article{kairouz2021advances,
  title={Advances and Open Problems in Federated Learning},
  author={Kairouz, Peter and McMahan, H Brendan and Avent, Brendan and Bellet, Aur{\'e}lien and Bennis, Mehdi and Bhagoji, Arjun Nitin and Bonawitz, Kallista and Charles, Zachary and Cormode, Graham and Cummings, Rachel and others},
  journal={Foundations and Trends{\textregistered} in Machine Learning},
  volume={14},
  number={1--2},
  pages={1--210},
  year={2021},
  publisher={Now Publishers, Inc.}
}

@inproceedings{hsieh2020non,
  title={The Non-IID Data Quagmire of Decentralized Machine Learning},
  author={Hsieh, Kevin and Phanishayee, Amar and Mutlu, Onur and Gibbons, Phillip},
  booktitle={International Conference on Machine Learning (ICML)},
  pages={4387--4398},
  year={2020},
  organization={PMLR}
}

@inproceedings{karimireddy2020scaffold,
  title={{SCAFFOLD}: Stochastic Controlled Averaging for Federated Learning},
  author={Karimireddy, Sai Praneeth and Kale, Satyen and Mohri, Mehryar and Reddi, Sashank and Stich, Sebastian and Suresh, Ananda Theertha},
  booktitle={International Conference on Machine Learning (ICML)},
  pages={5132--5143},
  year={2020},
  organization={PMLR}
}

@inproceedings{li2020federated,
  title={Federated Optimization in Heterogeneous Networks},
  author={Li, Tian and Sahu, Anit Kumar and Zaheer, Manzil and Sanjabi, Maziar and Talwalkar, Ameet and Smith, Virginia},
  booktitle={Proceedings of Machine Learning and Systems (MLSys)},
  volume={2},
  pages={429--450},
  year={2020}
}

@inproceedings{wang2020tackling,
  title={Tackling the Objective Inconsistency Problem in Heterogeneous Federated Optimization},
  author={Wang, Jianyu and Liu, Qinghua and Liang, Hao and Joshi, Gauri and Poor, H Vincent},
  booktitle={Advances in Neural Information Processing Systems (NeurIPS)},
  volume={33},
  pages={7611--7623},
  year={2020}
}

@inproceedings{li2021model,
  title={Model-Contrastive Federated Learning},
  author={Li, Qinbin and He, Bingsheng and Song, Dawn},
  booktitle={Proceedings of the IEEE/CVF Conference on Computer Vision and Pattern Recognition (CVPR)},
  pages={10713--10722},
  year={2021}
}

@article{olah2020zoom,
  title={Zoom In: An Introduction to Circuits},
  author={Olah, Chris and Cammarata, Nick and Schubert, Ludwig and Goh, Gabriel and Petrov, Michael and Carter, Shan},
  journal={Distill},
  volume={5},
  number={3},
  pages={e00024--001},
  year={2020},
  publisher={Distill}
}

@article{elhage2021mathematical,
  title={A Mathematical Framework for Transformer Circuits},
  author={Elhage, Nelson and Nanda, Neel and Olsson, Catherine and Henighan, Tom and Joseph, Nicholas and Mann, Ben and Askell, Amanda and Bai, Yuntao and Chen, Anna and Conerly, Tom and others},
  journal={Transformer Circuits Thread},
  year={2021}
}

@inproceedings{wang2022interpretability,
  title={Interpretability in the Wild: a Circuit for Indirect Object Identification in {GPT}-2 small},
  author={Wang, Kevin and Variengien, Alexandre and Conmy, Arthur and Shlegeris, Buck and Steinhardt, Jacob},
  booktitle={International Conference on Learning Representations (ICLR)},
  year={2023}
}

@inproceedings{conmy2023towards,
  title={Towards Automated Circuit Discovery for Mechanistic Interpretability},
  author={Conmy, Arthur and Mavor-Parker, Augustine N and Lynch, Aengus and Heimersheim, Stefan and Garriga-Alonso, Adri{\`a}},
  booktitle={Advances in Neural Information Processing Systems (NeurIPS)},
  volume={36},
  year={2023}
}

@inproceedings{alain2016understanding,
  title={Understanding intermediate layers using linear classifier probes},
  author={Alain, Guillaume and Bengio, Yoshua},
  booktitle={International Conference on Learning Representations (ICLR)},
  year={2017}
}

@article{bricken2023towards,
  title={Towards Monosemanticity: Decomposing Language Models With Dictionary Learning},
  author={Bricken, Trenton and Templeton, Adly and Batson, Joshua and Chen, Brian and Jermyn, Adam and Conerly, Tom and Turner, Nick and Anil, Cem and Denison, Carson and Askell, Amanda and others},
  journal={Transformer Circuits Thread},
  year={2023}
}

@inproceedings{cunningham2023sparse,
  title={Sparse Autoencoders Find Highly Interpretable Features in Language Models},
  author={Cunningham, Hoagy and Ewart, Aidan and Riggs, Logan and Huben, Robert and Sharkey, Lee},
  booktitle={International Conference on Learning Representations (ICLR)},
  year={2024}
}

@article{bengio2013estimating,
  title={Estimating or Propagating Gradients Through Stochastic Neurons for Conditional Computation},
  author={Bengio, Yoshua and L{\'e}onard, Nicholas and Courville, Aaron},
  journal={arXiv preprint arXiv:1308.3432},
  year={2013}
}

@inproceedings{li2021fedrs,
  title={{FedRS}: Federated Learning with Restricted Softmax for Label Distribution Non-IID Data},
  author={Li, Xin-Chun and Zhan, De-Chuan},
  booktitle={Proceedings of the 27th ACM SIGKDD Conference on Knowledge Discovery and Data Mining},
  pages={995--1005},
  year={2021},
  doi={10.1145/3447548.3467254}
}

@article{arivazhagan2019federated,
  title={Federated Learning with Personalization Layers},
  author={Arivazhagan, Manoj Ghuhan and Aggarwal, Vinay and Singh, Aaditya Kumar and Choudhary, Sunav},
  journal={arXiv preprint arXiv:1912.00818},
  year={2019}
}

@inproceedings{collins2021exploiting,
  title={Exploiting Shared Representations for Personalized Federated Learning},
  author={Collins, Liam and Hassani, Hamed and Mokhtari, Aryan and Shakkottai, Sanjay},
  booktitle={International Conference on Machine Learning},
  pages={2089--2099},
  year={2021},
  volume={139},
  series={Proceedings of Machine Learning Research},
  publisher={PMLR}
}

@inproceedings{oh2022fedbabu,
  title={{FedBABU}: Toward Enhanced Representation for Federated Image Classification},
  author={Oh, Jaehoon and Kim, SangMook and Yun, Se-Young},
  booktitle={International Conference on Learning Representations},
  year={2022},
  url={https://openreview.net/forum?id=HuaYQfggn5u}
}

@inproceedings{wortsman2022model,
  title={Model Soups: Averaging Weights of Multiple Fine-Tuned Models Improves Accuracy Without Increasing Inference Time},
  author={Wortsman, Mitchell and Ilharco, Gabriel and Gadre, Samir Yitzhak and Roelofs, Rebecca and Gontijo-Lopes, Raphael and Morcos, Ari S. and Namkoong, Hongseok and Farhadi, Ali and Carmon, Yair and Kornblith, Simon and Schmidt, Ludwig},
  booktitle={International Conference on Machine Learning},
  pages={23965--23998},
  year={2022},
  volume={162},
  series={Proceedings of Machine Learning Research},
  publisher={PMLR}
}

@inproceedings{matena2022merging,
  title={Merging Models with Fisher-Weighted Averaging},
  author={Matena, Michael S. and Raffel, Colin A.},
  booktitle={Advances in Neural Information Processing Systems},
  volume={35},
  year={2022}
}

@inproceedings{frankle2020linear,
  title={Linear Mode Connectivity and the Lottery Ticket Hypothesis},
  author={Frankle, Jonathan and Dziugaite, Gintare Karolina and Roy, Daniel M. and Carbin, Michael},
  booktitle={International Conference on Machine Learning},
  pages={3259--3269},
  year={2020},
  volume={119},
  series={Proceedings of Machine Learning Research},
  publisher={PMLR}
}

@inproceedings{ainsworth2023git,
  title={Git Re-Basin: Merging Models modulo Permutation Symmetries},
  author={Ainsworth, Samuel K. and Hayase, Jonathan and Srinivasa, Siddhartha},
  booktitle={International Conference on Learning Representations},
  year={2023},
  url={https://openreview.net/forum?id=CQsmMYmlP5T}
}

@inproceedings{ilharco2023editing,
  title={Editing Models with Task Arithmetic},
  author={Ilharco, Gabriel and Ribeiro, Marco Tulio and Wortsman, Mitchell and Schmidt, Ludwig and Hajishirzi, Hannaneh and Farhadi, Ali},
  booktitle={International Conference on Learning Representations},
  year={2023},
  url={https://openreview.net/forum?id=6t0Kwf8-jrj}
}

@inproceedings{yadav2023ties,
  title={{TIES}-Merging: Resolving Interference When Merging Models},
  author={Yadav, Prateek and Tam, Derek and Choshen, Leshem and Raffel, Colin A. and Bansal, Mohit},
  booktitle={Advances in Neural Information Processing Systems},
  volume={36},
  year={2023},
  url={https://proceedings.neurips.cc/paper_files/paper/2023/hash/1644c9af28ab7916874f6fd6228a9bcf-Abstract-Conference.html}
}

@inproceedings{hanna2024faithfulness,
  title={Have Faith in Faithfulness: Going Beyond Circuit Overlap When Finding Model Mechanisms},
  author={Hanna, Michael and Pezzelle, Sandro and Belinkov, Yonatan},
  booktitle={Mechanistic Interpretability Workshop at ICML},
  year={2024},
  url={https://openreview.net/forum?id=grXgesr5dT}
}

@inproceedings{raghu2017svcca,
  title={SVCCA: Singular Vector Canonical Correlation Analysis for Deep Learning Dynamics and Interpretability},
  author={Raghu, Maithra and Gilmer, Justin and Yosinski, Jason and Sohl-Dickstein, Jascha},
  booktitle={Advances in Neural Information Processing Systems (NeurIPS)},
  volume={30},
  year={2017}
}
\bibliographystyle{icml2026}

\newpage
\appendix
\onecolumn

\clearpage
\appendix

\section*{Appendix}
\phantomsection
\label{app:roadmap}

\vspace{0.75em}

\begin{center}
{\Large\bfseries Table of Contents}

\vspace{0.35em}

\begin{minipage}{0.92\textwidth}
\centering
\small
This appendix provides implementation details and additional results
supporting the analyses in the main text. Each entry links to the
corresponding appendix section.
\end{minipage}
\end{center}

\vspace{1.25em}

\noindent\begin{minipage}{\textwidth}
\large

\noindent
\hyperref[app:model_arch_extra]{\textbf{Appendix~\ref*{app:model_arch_extra}}\quad Model Architecture Details}
\dotfill
\hyperref[app:model_arch_extra]{\pageref*{app:model_arch_extra}}

\vspace{0.55em}

\noindent
\hyperref[app:circuit_extra]{\textbf{Appendix~\ref*{app:circuit_extra}}\quad Additional Circuit Similarity Results}
\dotfill
\hyperref[app:circuit_extra]{\pageref*{app:circuit_extra}}

\vspace{0.40em}

\noindent\hspace{2.0em}
\hyperref[app:circuit_suff_extra]{Appendix~\ref*{app:circuit_suff_extra}\quad Per-Class Circuit Sufficiency and Necessity}
\dotfill
\hyperref[app:circuit_suff_extra]{\pageref*{app:circuit_suff_extra}}

\vspace{0.55em}

\noindent
\hyperref[app:linear_probe_extra]{\textbf{Appendix~\ref*{app:linear_probe_extra}}\quad Additional Linear Probe Results}
\dotfill
\hyperref[app:linear_probe_extra]{\pageref*{app:linear_probe_extra}}

\vspace{0.55em}

\noindent
\hyperref[app:head_finetune_extra]{\textbf{Appendix~\ref*{app:head_finetune_extra}}\quad Additional Head-Only Finetuning Results}
\dotfill
\hyperref[app:head_finetune_extra]{\pageref*{app:head_finetune_extra}}

\vspace{0.55em}

\noindent
\hyperref[app:usae_extra]{\textbf{Appendix~\ref*{app:usae_extra}}\quad Additional USAE Results}
\dotfill
\hyperref[app:usae_extra]{\pageref*{app:usae_extra}}

\vspace{0.55em}

\noindent
\hyperref[app:svcca_alignment]{\textbf{Appendix~\ref*{app:svcca_alignment}}\quad Feature Alignment via SVCCA Intervention}
\dotfill
\hyperref[app:svcca_alignment]{\pageref*{app:svcca_alignment}}

\vspace{0.55em}

\noindent
\hyperref[app:dense_model_justification]{\textbf{Appendix~\ref*{app:dense_model_justification}}\quad Generalization to Dense Models and Justification for Sparse Training}
\dotfill
\hyperref[app:dense_model_justification]{\pageref*{app:dense_model_justification}}

\end{minipage}

\vfill

\noindent{\small\textit{Note:} Appendix~\ref{app:circuit_extra} contains the additional circuit-overlap figures and per-class sufficiency and necessity measurements for the remaining dataset--backbone pairs. Appendices~\ref{app:linear_probe_extra}--\ref{app:usae_extra} report the corresponding probing, finetuning, and concept-transfer results. Appendix~\ref{app:dense_model_justification} provides the dense-model control experiments confirming that our findings are not artifacts of sparse training.}

\clearpage

\section{Model Architecture Details}
\label{app:model_arch_extra}

Table~\ref{tab:model_architectures} summarizes the layer structure of the two architectures used in our experiments. Output sizes are shown for CIFAR-10 inputs ($3\times32\times32$). For Fashion-MNIST, the same architectures are used with single-channel input; the corresponding final spatial sizes are $3\times3$ for the CNN and $1\times1$ for the ResNet. Circuit discovery gates convolutional layers but excludes the final classifier and ResNet downsample projections.

\begin{table*}[t]
  \caption{Model architectures used in our experiments. Output sizes are shown for CIFAR-10 inputs ($3\times32\times32$). For Fashion-MNIST, the same architectures are used with single-channel input; the corresponding final spatial sizes are $3\times3$ for the CNN and $1\times1$ for the ResNet. Circuit discovery gates convolutional layers but excludes the final classifier and ResNet downsample projections.}
  \label{tab:model_architectures}
  \centering
  \begin{small}
  \begin{sc}
  \begin{tabular}{llcccc}
    \toprule
    Model & Layer & Channels & Kernel / Stride / Padding & Output Size & Gated? \\
    \midrule
    \multirow{4}{*}{CNN}
      & conv1 & $3 \rightarrow 64$ & $3 / 1 / 1$ & $64 \times 16 \times 16$ & Yes \\
      & conv2 & $64 \rightarrow 128$ & $3 / 1 / 1$ & $128 \times 8 \times 8$ & Yes \\
      & conv3 & $128 \rightarrow 256$ & $3 / 1 / 1$ & $256 \times 4 \times 4$ & Yes \\
      & fc & $4096 \rightarrow 10$ & -- & $10$ & No \\
    \midrule
    \multirow{10}{*}{ResNet}
      & conv1 & $3 \rightarrow 64$ & $7 / 2 / 3$ & $64 \times 8 \times 8$ & Yes \\
      & block1.0.conv1 & $64 \rightarrow 64$ & $3 / 1 / 1$ & $64 \times 8 \times 8$ & Yes \\
      & block1.0.conv2 & $64 \rightarrow 64$ & $3 / 1 / 1$ & $64 \times 8 \times 8$ & Yes \\
      & block2.0.conv1 & $64 \rightarrow 128$ & $3 / 2 / 1$ & $128 \times 4 \times 4$ & Yes \\
      & block2.0.conv2 & $128 \rightarrow 128$ & $3 / 1 / 1$ & $128 \times 4 \times 4$ & Yes \\
      & block3.0.conv1 & $128 \rightarrow 192$ & $3 / 2 / 1$ & $192 \times 2 \times 2$ & Yes \\
      & block3.0.conv2 & $192 \rightarrow 192$ & $3 / 1 / 1$ & $192 \times 2 \times 2$ & Yes \\
      & block4.0.conv1 & $192 \rightarrow 256$ & $3 / 2 / 1$ & $256 \times 1 \times 1$ & Yes \\
      & block4.0.conv2 & $256 \rightarrow 256$ & $3 / 1 / 1$ & $256 \times 1 \times 1$ & Yes \\
      & fc & $256 \rightarrow 10$ & -- & $10$ & No \\
    \bottomrule
  \end{tabular}
  \end{sc}
  \end{small}
\end{table*}

\clearpage

\section{Additional Circuit Similarity Results}
\label{app:circuit_extra}

\begin{figure}[htbp]
  \centering
  \begin{subfigure}{0.32\linewidth}
    \centering
    \includegraphics[width=\linewidth]{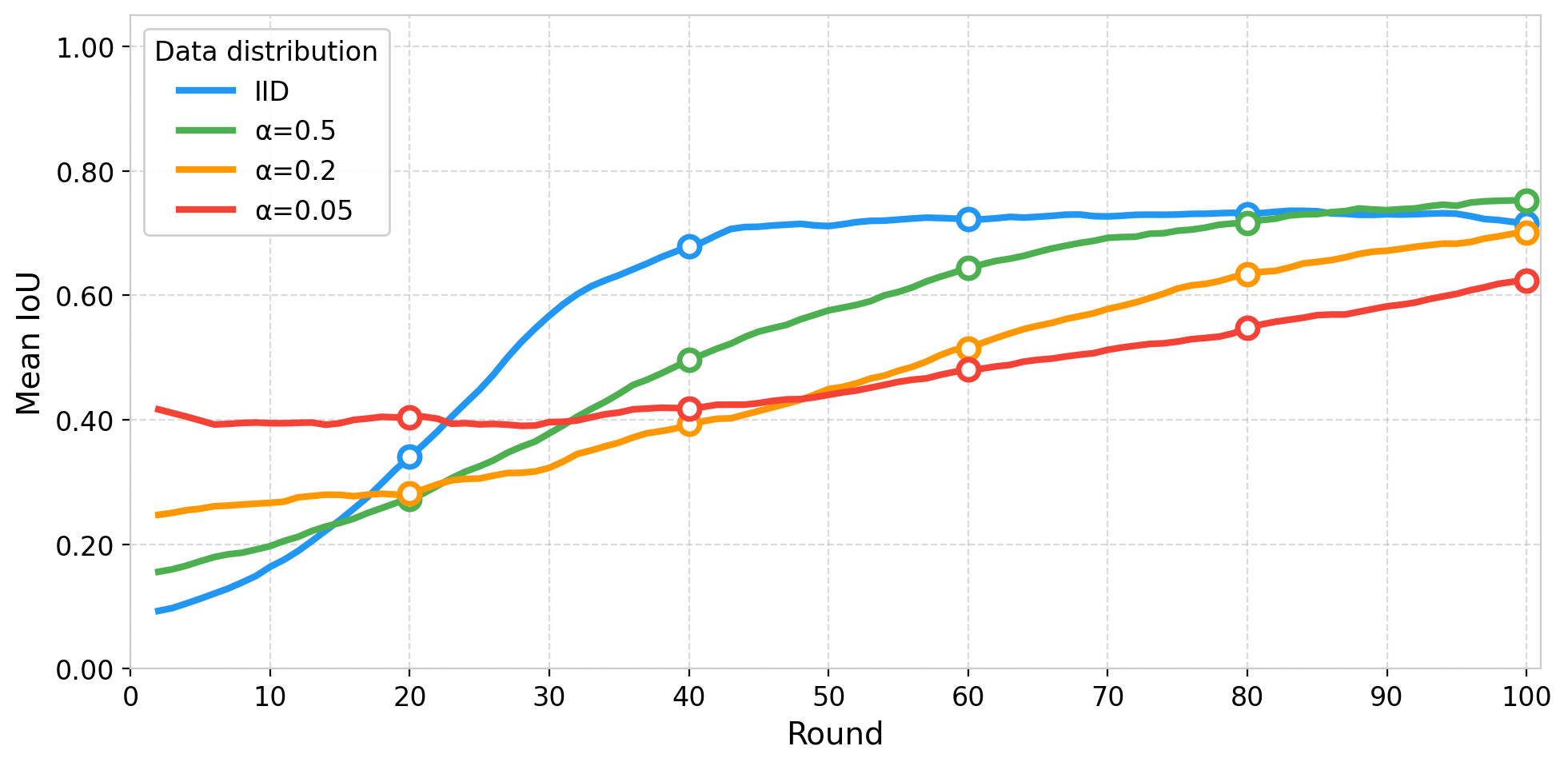}
    \caption{Intra-client stability.}
    \label{fig:cifar_resnet_intra_appendix}
  \end{subfigure}
  \hfill
  \begin{subfigure}{0.32\linewidth}
    \centering
    \includegraphics[width=\linewidth]{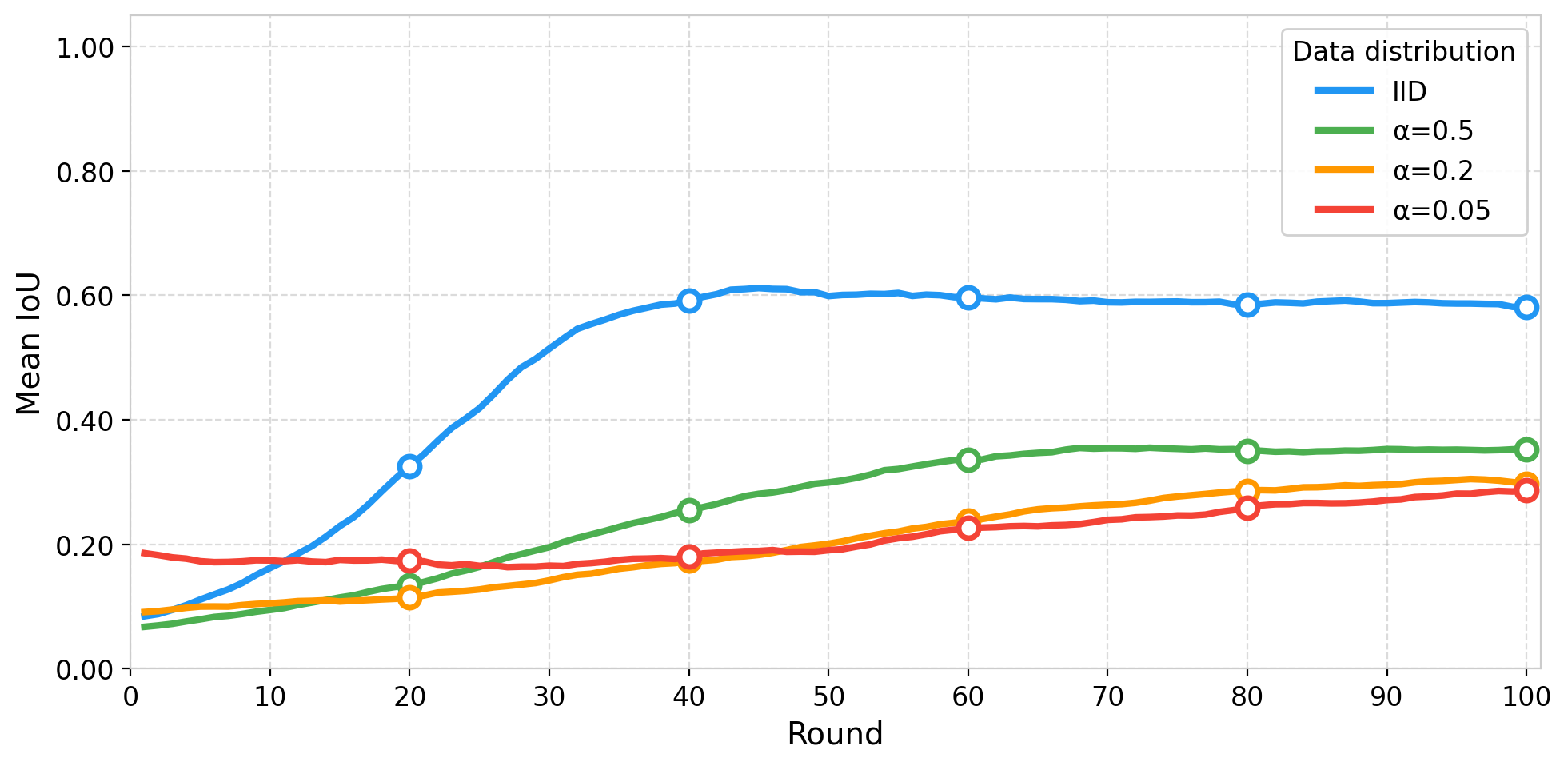}
    \caption{Inter-client consistency.}
    \label{fig:cifar_resnet_inter_client_appendix}
  \end{subfigure}
  \hfill
  \begin{subfigure}{0.32\linewidth}
    \centering
    \includegraphics[width=\linewidth]{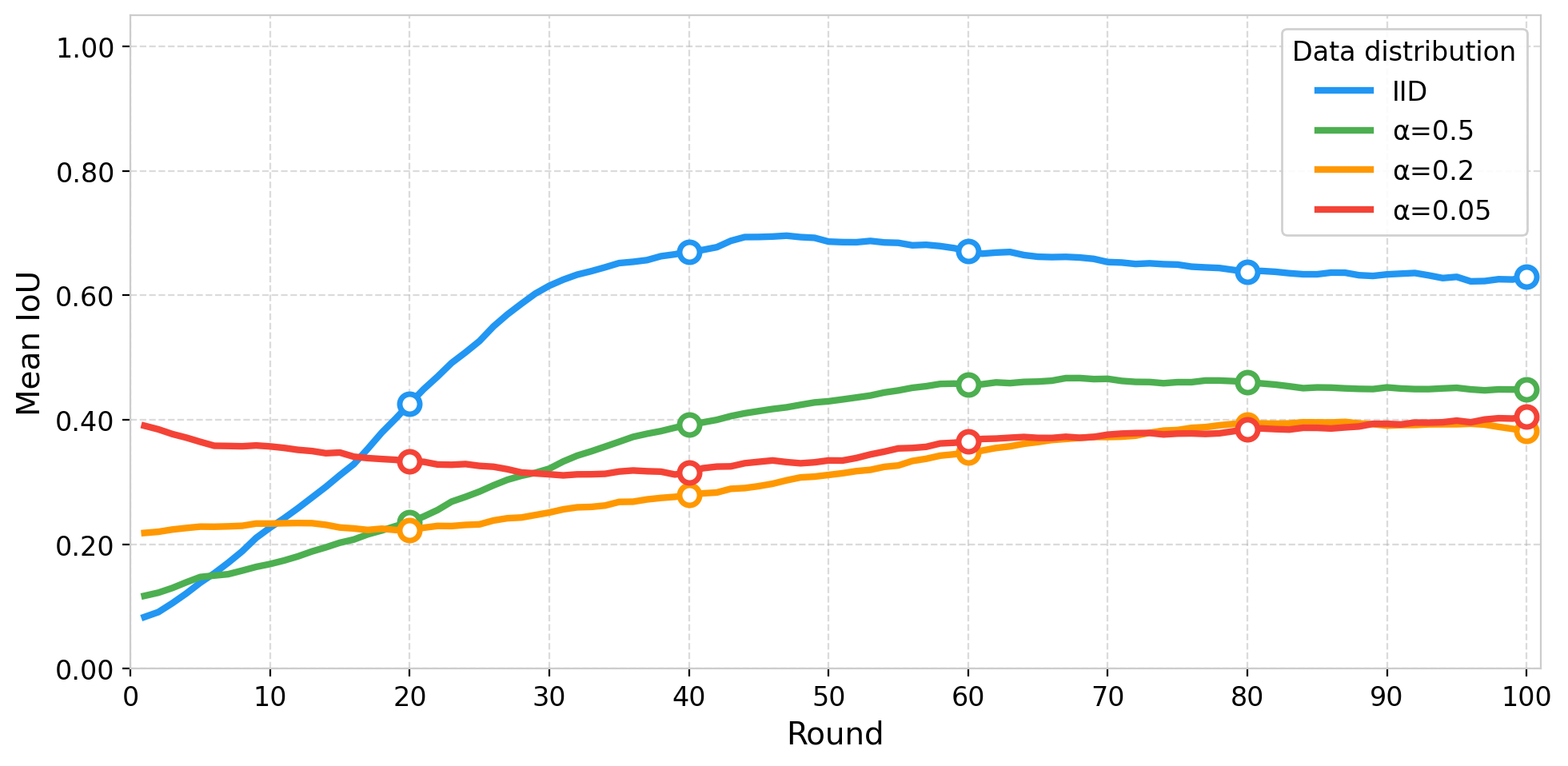}
    \caption{Local-to-global preservation.}
    \label{fig:cifar_resnet_local_global_appendix}
  \end{subfigure}
  \caption{Circuit similarity trends on CIFAR-10 with ResNet backbone.}
  \label{fig:cifar_resnet_circuit_appendix}
\end{figure}

\begin{figure}[htbp]
  \centering
  \begin{subfigure}{0.32\linewidth}
    \centering
    \includegraphics[width=\linewidth]{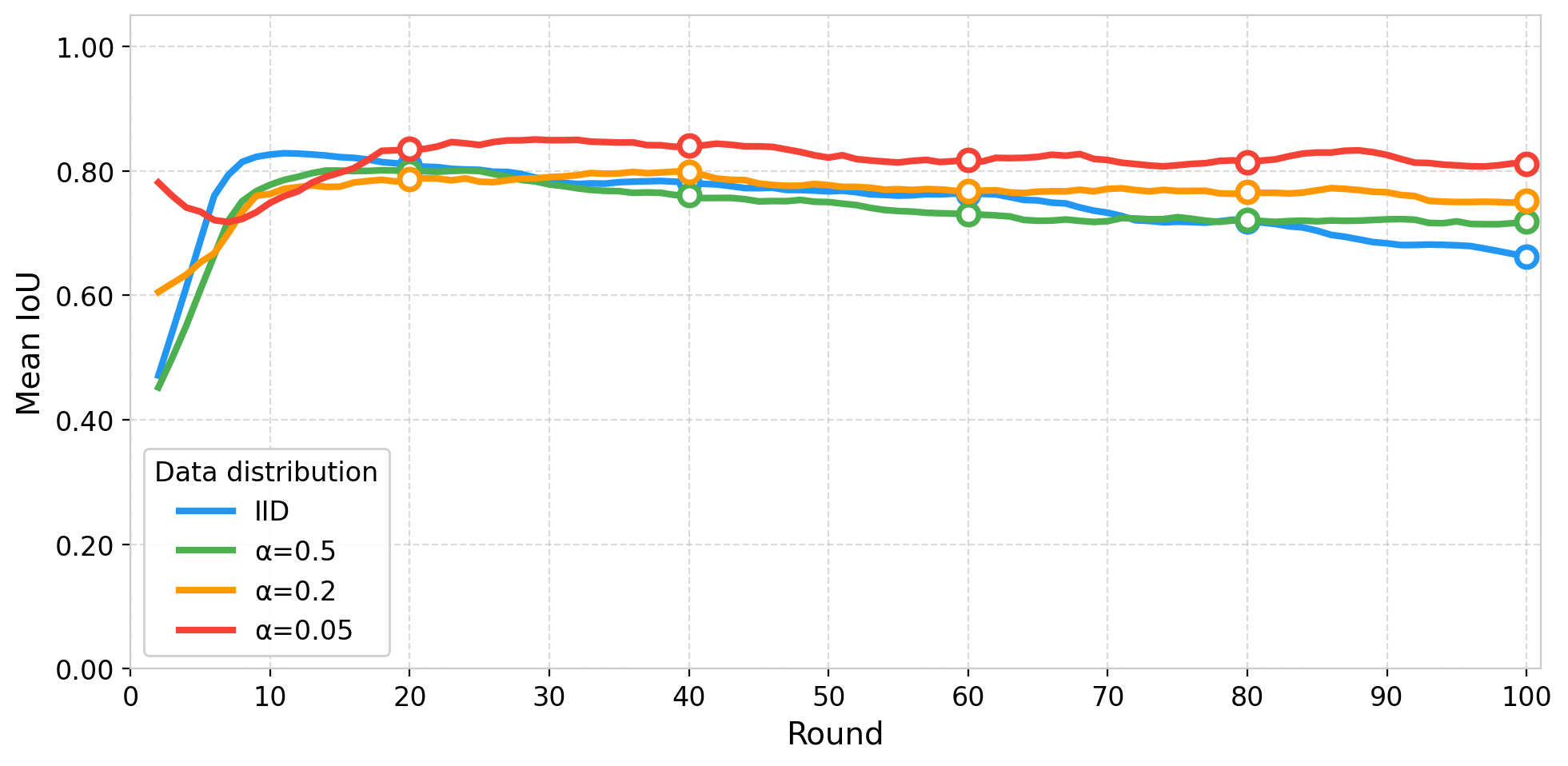}
    \caption{Intra-client stability.}
    \label{fig:fmnist_cnn_intra_appendix}
  \end{subfigure}
  \hfill
  \begin{subfigure}{0.32\linewidth}
    \centering
    \includegraphics[width=\linewidth]{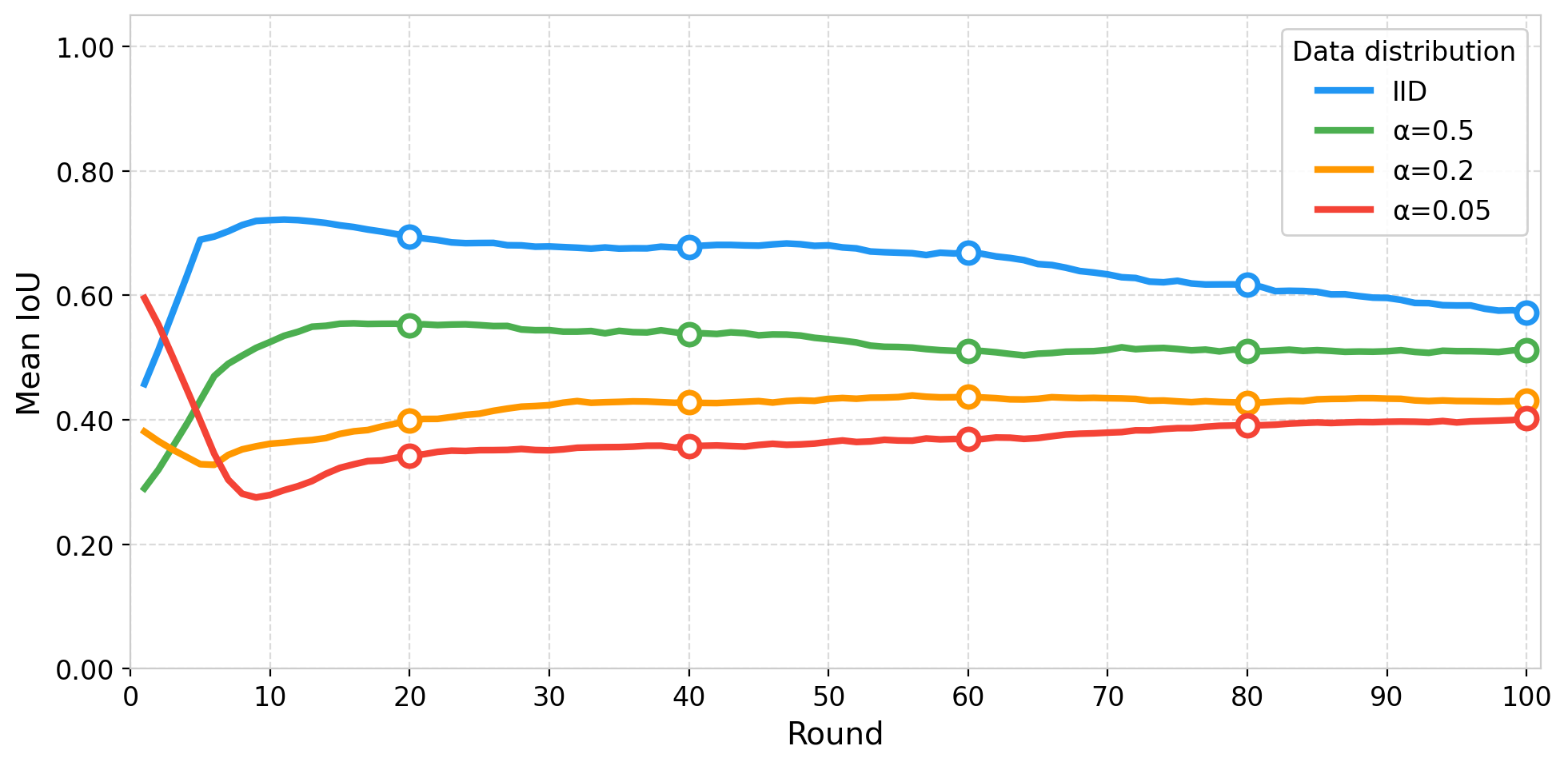}
    \caption{Inter-client consistency.}
    \label{fig:fmnist_cnn_inter_client_appendix}
  \end{subfigure}
  \hfill
  \begin{subfigure}{0.32\linewidth}
    \centering
    \includegraphics[width=\linewidth]{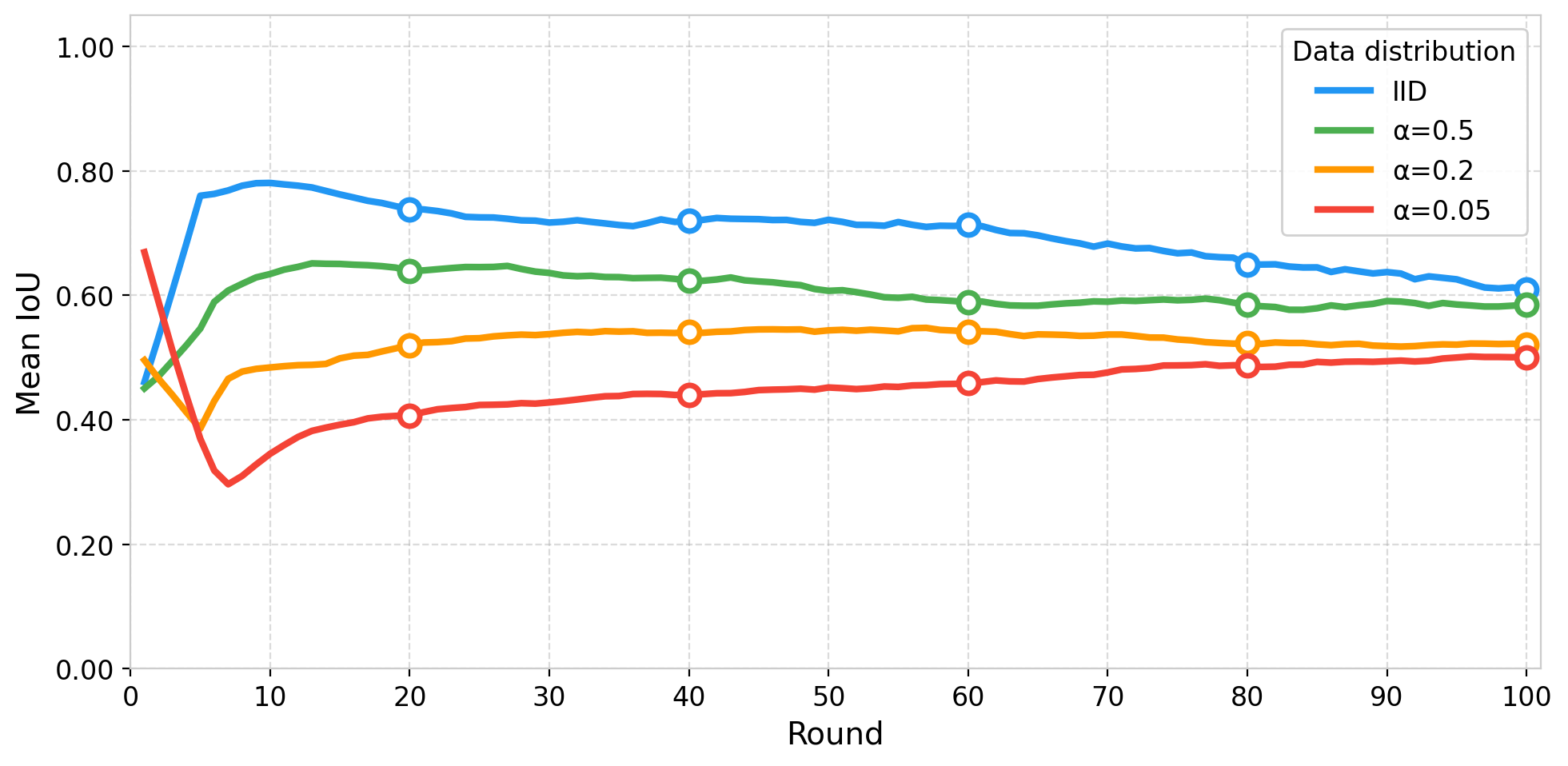}
    \caption{Local-to-global preservation.}
    \label{fig:fmnist_cnn_local_global_appendix}
  \end{subfigure}
  \caption{Circuit similarity trends on Fashion-MNIST with CNN backbone.}
  \label{fig:fmnist_cnn_circuit_appendix}
\end{figure}

\begin{figure}[htbp]
  \centering
  \begin{subfigure}{0.32\linewidth}
    \centering
    \includegraphics[width=\linewidth]{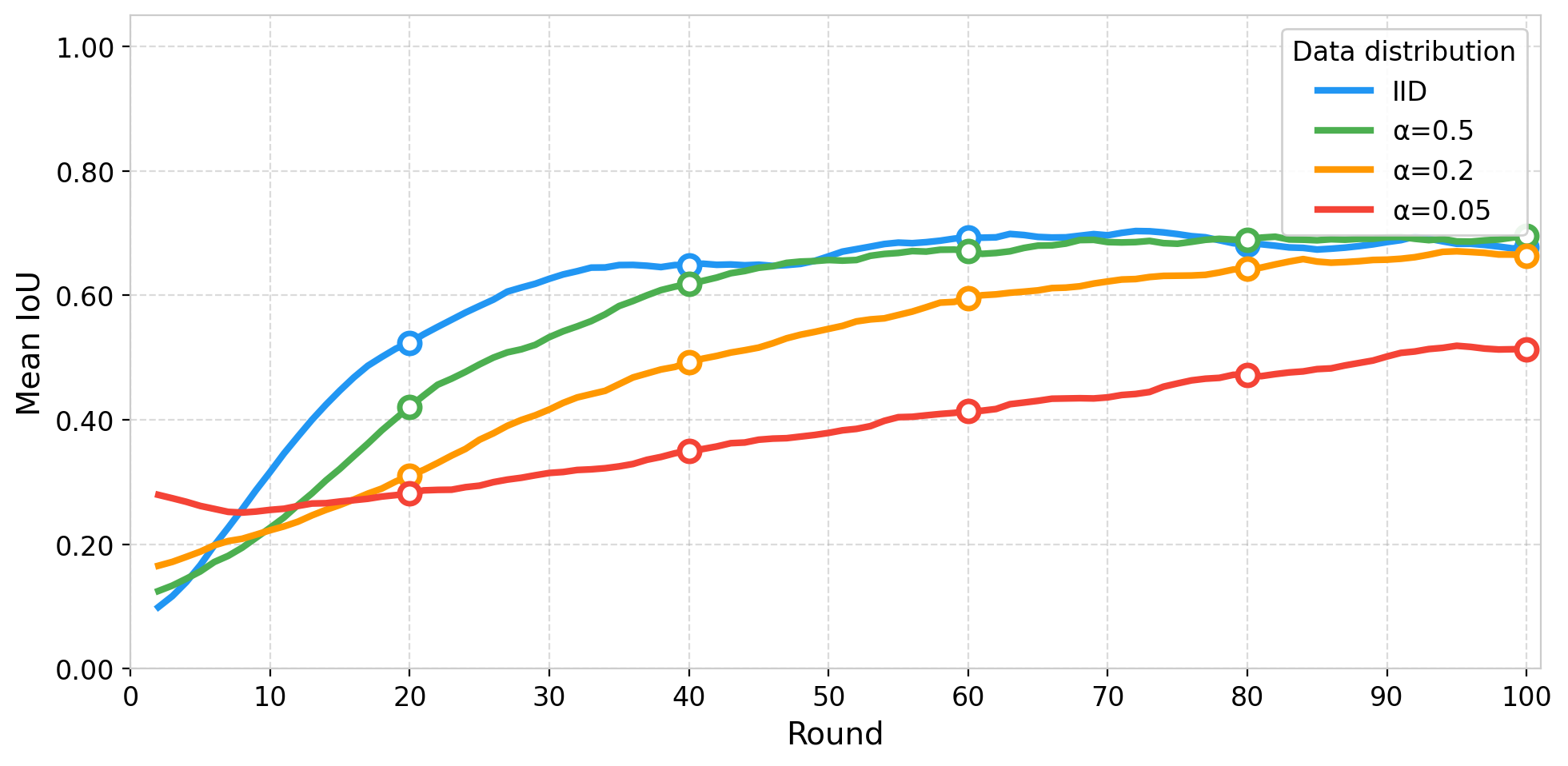}
    \caption{Intra-client stability.}
    \label{fig:fmnist_resnet_intra_appendix}
  \end{subfigure}
  \hfill
  \begin{subfigure}{0.32\linewidth}
    \centering
    \includegraphics[width=\linewidth]{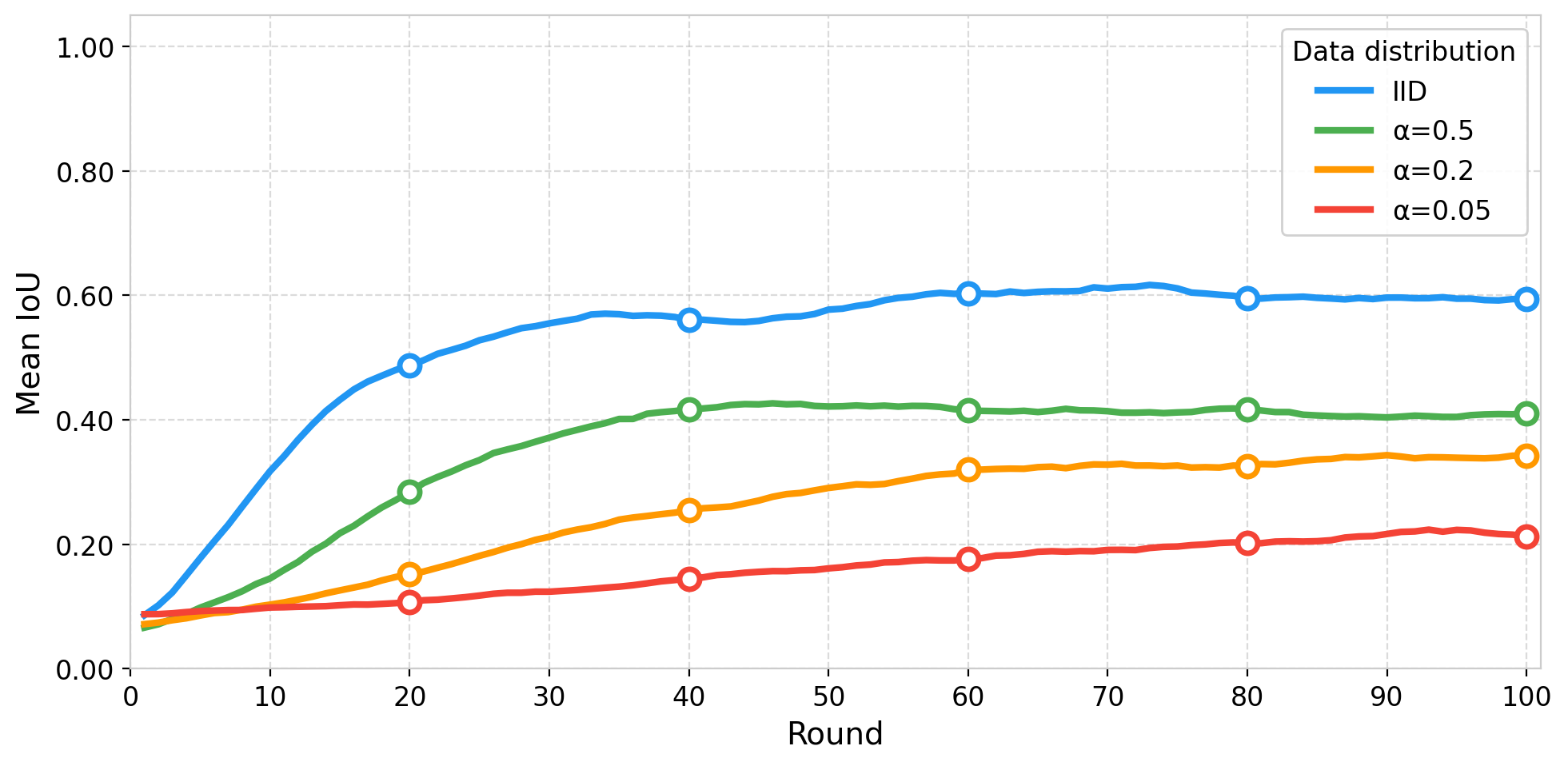}
    \caption{Inter-client consistency.}
    \label{fig:fmnist_resnet_inter_client_appendix}
  \end{subfigure}
  \hfill
  \begin{subfigure}{0.32\linewidth}
    \centering
    \includegraphics[width=\linewidth]{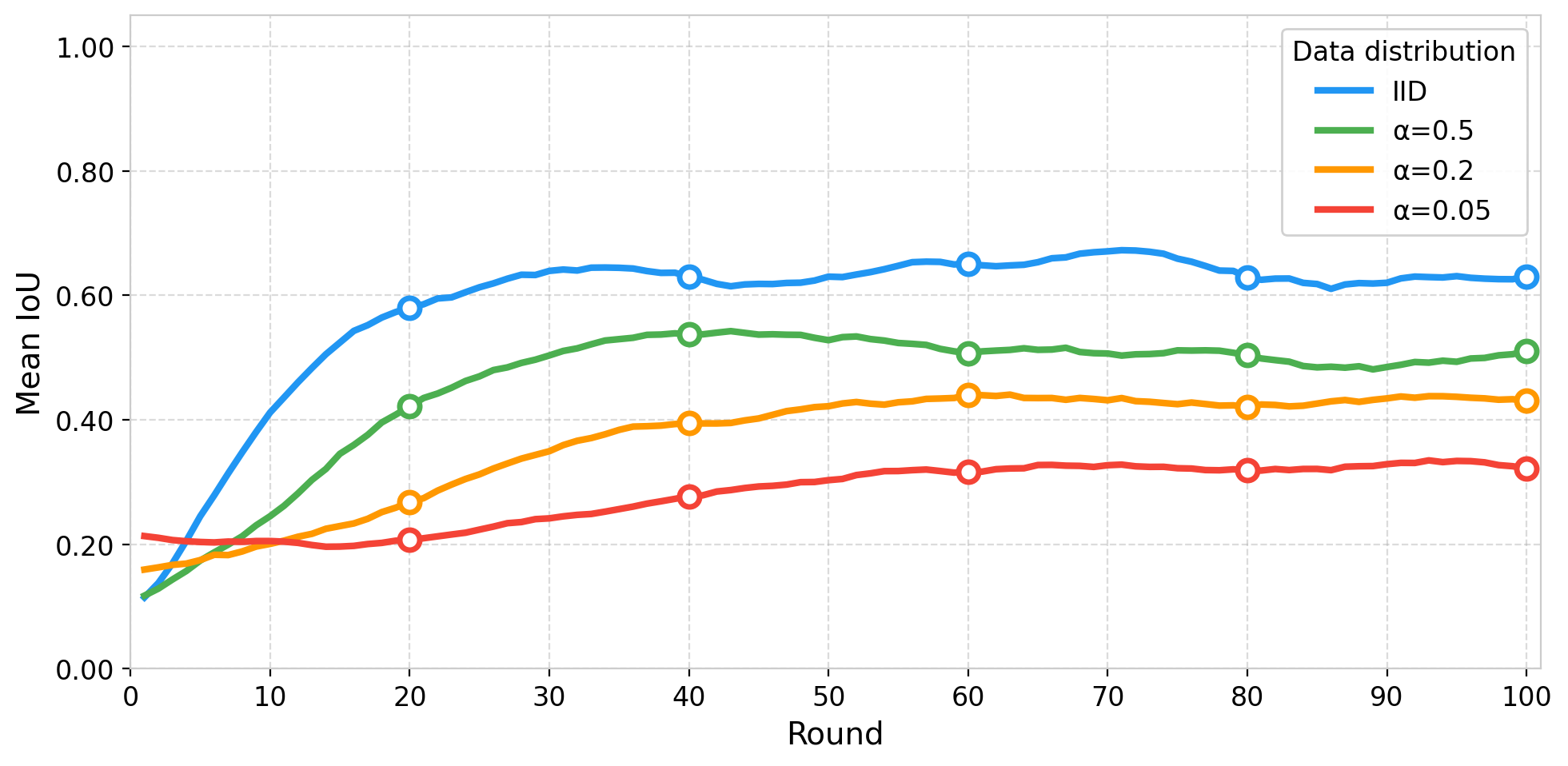}
    \caption{Local-to-global preservation.}
    \label{fig:fmnist_resnet_local_global_appendix}
  \end{subfigure}
  \caption{Circuit similarity trends on Fashion-MNIST with ResNet backbone.}
  \label{fig:fmnist_resnet_circuit_appendix}
\end{figure}

Figures~\ref{fig:cifar_resnet_circuit_appendix}--\ref{fig:fmnist_resnet_circuit_appendix} show intra-client circuit stability, inter-client circuit consistency, and local-to-global circuit preservation for the remaining dataset and architecture pairs. Each figure shows all three metrics together for a given dataset--backbone combination.

\subsection{Per-Class Circuit Sufficiency and Necessity}
\label{app:circuit_suff_extra}

Tables~\ref{tab:circuit_cifar_resnet_005}--\ref{tab:circuit_fmnist_resnet_005} extend the per-class circuit sufficiency and necessity analysis from Table~\ref{tab:per_class_acc} to the remaining dataset and backbone pairs under $\alpha=0.05$. Circuit sufficiency is the target-class sufficiency score of the extracted class-specific circuit; necessity measures the remaining accuracy when the circuit is ablated; nodes denotes the number of active gated channels in the extracted circuit.

Notably, we observe a distinct cross-family difference in these results: CIFAR-10 circuits are almost universally necessary (ablation degrades accuracy to near $0\%$), with only the Bird class acting as an exception across both architectures due to backup pathways. In contrast, Fashion-MNIST displays the opposite trend for its best-performing classes. High-accuracy classes like Bag, Ankle boot, and Dress retain substantial accuracy after their main circuits are ablated, suggesting the model learned distributed, redundant representations rather than purely sparse pathways for those concepts under non-IID training. This discrepancy is likely due to the relative simplicity of the Fashion-MNIST task; because the features are comparatively easy to learn, the models have ample capacity to develop multiple, overlapping backup pathways. Consequently, even when the primary class-specific circuit is ablated, these secondary pathways can easily compensate and successfully classify the inputs.

\begin{table}[htbp]
  \caption{Per-class global accuracy, circuit sufficiency, and necessity for CIFAR-10
  with ResNet backbone under $\alpha=0.05$. Overall global accuracy is
  34.18\%.}
  \label{tab:circuit_cifar_resnet_005}
  \centering
  \begin{small}
  \begin{sc}
  \begin{tabular}{lcccc}
    \toprule
    Class & Global Acc & Circuit Suff. & Necessity & Nodes \\
    \midrule
    Airplane   & 10.8\% & 100.0\% &  0.0\% & 66 \\
    Automobile & 27.0\% & 100.0\% &  0.0\% & 32 \\
    Bird       & 79.4\% & 100.0\% & 17.3\% & 24 \\
    Cat        &  0.0\% &  97.5\% &  0.0\% & 88 \\
    Deer       &  0.5\% &  99.9\% &  0.0\% & 66 \\
    Dog        & 43.0\% & 100.0\% &  0.3\% & 23 \\
    Frog       & 44.3\% & 100.0\% &  0.0\% & 37 \\
    Horse      &  7.9\% & 100.0\% &  0.0\% & 51 \\
    Ship       & 52.8\% & 100.0\% &  0.0\% & 44 \\
    Truck      & 76.1\% & 100.0\% &  0.0\% & 32 \\
    \bottomrule
  \end{tabular}
  \end{sc}
  \end{small}
\end{table}

\begin{table}[htbp]
  \caption{Per-class global accuracy, circuit sufficiency, and necessity for
  Fashion-MNIST with CNN backbone under $\alpha=0.05$. Overall global
  accuracy is 76.74\%.}
  \label{tab:circuit_fmnist_cnn_005}
  \centering
  \begin{small}
  \begin{sc}
  \begin{tabular}{lcccc}
    \toprule
    Class & Global Acc & Circuit Suff. & Necessity & Nodes \\
    \midrule
    T-shirt    & 42.1\% &  98.5\% &  0.0\% & 47 \\
    Trouser    & 93.3\% & 100.0\% &  0.0\% & 19 \\
    Pullover   & 52.0\% &  99.8\% &  0.0\% & 40 \\
    Dress      & 86.8\% &  99.3\% & 34.0\% & 48 \\
    Coat       & 44.6\% &  99.9\% &  0.0\% & 46 \\
    Sandal     & 74.6\% & 100.0\% &  0.4\% & 47 \\
    Shirt      & 92.0\% &  99.8\% & 49.1\% & 38 \\
    Sneaker    & 87.0\% & 100.0\% &  0.1\% & 26 \\
    Bag        & 96.5\% &  99.4\% & 81.3\% & 30 \\
    Ankle boot & 98.5\% & 100.0\% & 85.8\% & 26 \\
    \bottomrule
  \end{tabular}
  \end{sc}
  \end{small}
\end{table}

\begin{table}[htbp]
  \caption{Per-class global accuracy, circuit sufficiency, and necessity for
  Fashion-MNIST with ResNet backbone under $\alpha=0.05$. Overall global
  accuracy is 72.39\%.}
  \label{tab:circuit_fmnist_resnet_005}
  \centering
  \begin{small}
  \begin{sc}
  \begin{tabular}{lcccc}
    \toprule
    Class & Global Acc & Circuit Suff. & Necessity & Nodes \\
    \midrule
    T-shirt    & 65.7\% & 100.0\% &  0.0\% & 36 \\
    Trouser    & 91.6\% &  99.8\% &  0.0\% & 36 \\
    Pullover   & 64.5\% & 100.0\% &  0.0\% & 42 \\
    Dress      & 80.3\% & 100.0\% & 38.7\% & 25 \\
    Coat       & 40.7\% &  99.6\% &  0.0\% & 53 \\
    Sandal     & 63.8\% & 100.0\% &  0.0\% & 46 \\
    Shirt      & 76.3\% & 100.0\% &  0.1\% & 32 \\
    Sneaker    & 45.1\% & 100.0\% &  0.0\% & 41 \\
    Bag        & 96.7\% & 100.0\% & 58.8\% & 12 \\
    Ankle boot & 99.2\% & 100.0\% &  0.2\% & 19 \\
    \bottomrule
  \end{tabular}
  \end{sc}
  \end{small}
\end{table}

\clearpage

\section{Additional Linear Probe Results}
\label{app:linear_probe_extra}

Table~\ref{tab:linear_probe_cifar_resnet} reports linear probe results for CIFAR-10 with a ResNet backbone. Tables~\ref{tab:linear_probe_fmnist_cnn} and~\ref{tab:linear_probe_fmnist_resnet} report the corresponding Fashion-MNIST results.

\begin{table}[htbp]
  \caption{Global model accuracy vs.\ linear probe accuracy on CIFAR-10 with ResNet backbone.
  Same probe setup as Table~\ref{tab:linear_probe}.}
  \label{tab:linear_probe_cifar_resnet}
  \centering
  \begin{small}
  \begin{sc}
  \begin{tabular}{lccc}
    \toprule
    Config & Global Acc & Probe Acc & Delta \\
    \midrule
    IID           & 59.7\% & 54.0\% & $-$5.7\% \\
    $\alpha=0.5$  & 56.8\% & 53.5\% & $-$3.3\% \\
    $\alpha=0.2$  & 49.7\% & 50.1\% & +0.4\%   \\
    $\alpha=0.05$ & 34.2\% & 49.4\% & +15.2\%  \\
    \bottomrule
  \end{tabular}
  \end{sc}
  \end{small}
\end{table}

\begin{table}[htbp]
  \caption{Global model accuracy vs.\ linear probe accuracy on Fashion-MNIST with CNN backbone.
  Same probe setup as Table~\ref{tab:linear_probe}.}
  \label{tab:linear_probe_fmnist_cnn}
  \centering
  \begin{small}
  \begin{sc}
  \begin{tabular}{lccc}
    \toprule
    Config & Global Acc & Probe Acc & Delta \\
    \midrule
    IID           & 90.9\% & 91.2\% & +0.3\%  \\
    $\alpha=0.5$  & 90.0\% & 90.8\% & +0.8\%  \\
    $\alpha=0.2$  & 88.8\% & 90.3\% & +1.5\%  \\
    $\alpha=0.05$ & 76.7\% & 89.8\% & +13.1\% \\
    \bottomrule
  \end{tabular}
  \end{sc}
  \end{small}
\end{table}

\begin{table}[htbp]
  \caption{Global model accuracy vs.\ linear probe accuracy on Fashion-MNIST with ResNet backbone.
  Same probe setup as Table~\ref{tab:linear_probe}.}
  \label{tab:linear_probe_fmnist_resnet}
  \centering
  \begin{small}
  \begin{sc}
  \begin{tabular}{lccc}
    \toprule
    Config & Global Acc & Probe Acc & Delta \\
    \midrule
    IID           & 89.7\% & 86.5\% & $-$3.3\%  \\
    $\alpha=0.5$  & 88.5\% & 86.8\% & $-$1.7\%  \\
    $\alpha=0.2$  & 87.1\% & 85.7\% & $-$1.4\%  \\
    $\alpha=0.05$ & 72.4\% & 83.4\% & +11.0\% \\
    \bottomrule
  \end{tabular}
  \end{sc}
  \end{small}
\end{table}

While the end-to-end trained head slightly outperforms the linear probe in the balanced IID ResNet10 case, the critical finding is the significant shift in the accuracy delta from $-$5.7\% (IID) to +15.2\% ($\alpha=0.05$). This shows that discriminative features remain accessible in the backbone even as the default prediction pathway fails.

\clearpage

\section{Additional Head-Only Finetuning Results}
\label{app:head_finetune_extra}

Tables~\ref{tab:head_finetune_cifar_resnet}--\ref{tab:head_finetune_fmnist_resnet} report head-only finetuning results for the additional dataset and backbone pairs. As in Table~\ref{tab:head_finetune}, the backbone is frozen throughout and only the final classification head is updated on a small balanced IID subset.

\begin{table}[htbp]
  \caption{Head-only finetuning on CIFAR-10 with ResNet backbone.
  The backbone is frozen and only the classification head is finetuned on
  200 balanced IID samples for 5 epochs.}
  \label{tab:head_finetune_cifar_resnet}
  \centering
  \begin{small}
  \begin{sc}
  \begin{tabular}{lccc}
    \toprule
    Config & Pre-FT & Post-FT & Delta \\
    \midrule
    IID           & 59.7\% & 59.4\% & $-0.3\%$ \\
    $\alpha=0.5$  & 56.8\% & 56.6\% & $-0.3\%$ \\
    $\alpha=0.2$  & 49.7\% & 52.2\% & $+2.4\%$ \\
    $\alpha=0.05$ & 34.2\% & 42.9\% & $+8.7\%$ \\
    \bottomrule
  \end{tabular}
  \end{sc}
  \end{small}
\end{table}

\begin{table}[htbp]
  \caption{Head-only finetuning on Fashion-MNIST with CNN backbone.
  The backbone is frozen and only the classification head is finetuned on
  200 balanced IID samples for 5 epochs.}
  \label{tab:head_finetune_fmnist_cnn}
  \centering
  \begin{small}
  \begin{sc}
  \begin{tabular}{lccc}
    \toprule
    Config & Pre-FT & Post-FT & Delta \\
    \midrule
    IID           & 90.9\% & 90.1\% & $-0.8\%$ \\
    $\alpha=0.5$  & 90.0\% & 89.6\% & $-0.4\%$ \\
    $\alpha=0.2$  & 88.8\% & 88.5\% & $-0.2\%$ \\
    $\alpha=0.05$ & 76.7\% & 86.3\% & $+9.6\%$ \\
    \bottomrule
  \end{tabular}
  \end{sc}
  \end{small}
\end{table}

\begin{table}[htbp]
  \caption{Head-only finetuning on Fashion-MNIST with ResNet backbone.
  The backbone is frozen and only the classification head is finetuned on
  200 balanced IID samples for 5 epochs.}
  \label{tab:head_finetune_fmnist_resnet}
  \centering
  \begin{small}
  \begin{sc}
  \begin{tabular}{lccc}
    \toprule
    Config & Pre-FT & Post-FT & Delta \\
    \midrule
    IID           & 89.7\% & 89.6\% & $-0.2\%$ \\
    $\alpha=0.5$  & 88.5\% & 88.6\% & $+0.1\%$ \\
    $\alpha=0.2$  & 87.1\% & 87.4\% & $+0.3\%$ \\
    $\alpha=0.05$ & 72.4\% & 82.2\% & $+9.9\%$ \\
    \bottomrule
  \end{tabular}
  \end{sc}
  \end{small}
\end{table}

\clearpage

\section{Additional USAE Results}
\label{app:usae_extra}

Tables~\ref{tab:usae_cifar_resnet}--\ref{tab:usae_fmnist_resnet} report USAE concept-transfer results for the additional dataset and backbone pairs. Same-model entries are reconstruction upper bounds, while cross-model entries test transfer between IID and non-IID activation spaces.

\begin{table}[htbp]
  \caption{USAE concept-transfer accuracy on CIFAR-10 with ResNet backbone.
  Same-model entries are reconstruction upper bounds;
  cross-model entries test concept transfer.}
  \label{tab:usae_cifar_resnet}
  \centering
  \begin{small}
  \begin{sc}
  \begin{tabular}{lcc}
    \toprule
    Dictionary & Source $\to$ Target & Accuracy \\
    \midrule
    Shared (1868)
        & IID $\to$ IID         & 0.5830 \\
        & IID $\to$ Non-IID     & 0.3312 \\
        & Non-IID $\to$ IID     & 0.5533 \\
        & Non-IID $\to$ Non-IID & 0.3454 \\
    \midrule
    Full (2048)
        & IID $\to$ IID         & 0.5909 \\
        & IID $\to$ Non-IID     & 0.3353 \\
        & Non-IID $\to$ IID     & 0.5533 \\
        & Non-IID $\to$ Non-IID & 0.3454 \\
    \bottomrule
  \end{tabular}
  \end{sc}
  \end{small}
\end{table}

\begin{table}[htbp]
  \caption{USAE concept-transfer accuracy on Fashion-MNIST with CNN backbone.
  Same-model entries are reconstruction upper bounds;
  cross-model entries test concept transfer.}
  \label{tab:usae_fmnist_cnn}
  \centering
  \begin{small}
  \begin{sc}
  \begin{tabular}{lcc}
    \toprule
    Dictionary & Source $\to$ Target & Accuracy \\
    \midrule
    Shared (1985)
        & IID $\to$ IID         & 0.8865 \\
        & IID $\to$ Non-IID     & 0.7344 \\
        & Non-IID $\to$ IID     & 0.8761 \\
        & Non-IID $\to$ Non-IID & 0.7627 \\
    \midrule
    Full (2048)
        & IID $\to$ IID         & 0.8889 \\
        & IID $\to$ Non-IID     & 0.7361 \\
        & Non-IID $\to$ IID     & 0.8774 \\
        & Non-IID $\to$ Non-IID & 0.7566 \\
    \bottomrule
  \end{tabular}
  \end{sc}
  \end{small}
\end{table}

\begin{table}[htbp]
  \caption{USAE concept-transfer accuracy on Fashion-MNIST with ResNet backbone.
  Same-model entries are reconstruction upper bounds;
  cross-model entries test concept transfer.}
  \label{tab:usae_fmnist_resnet}
  \centering
  \begin{small}
  \begin{sc}
  \begin{tabular}{lcc}
    \toprule
    Dictionary & Source $\to$ Target & Accuracy \\
    \midrule
    Shared (1864)
        & IID $\to$ IID         & 0.8910 \\
        & IID $\to$ Non-IID     & 0.7048 \\
        & Non-IID $\to$ IID     & 0.8711 \\
        & Non-IID $\to$ Non-IID & 0.7263 \\
    \midrule
    Full (2048)
        & IID $\to$ IID         & 0.8938 \\
        & IID $\to$ Non-IID     & 0.7323 \\
        & Non-IID $\to$ IID     & 0.8729 \\
        & Non-IID $\to$ Non-IID & 0.7262 \\
    \bottomrule
  \end{tabular}
  \end{sc}
  \end{small}
\end{table}
\clearpage

\section{Feature Alignment via SVCCA Intervention}
\label{app:svcca_alignment}

To further investigate the geometric nature of representation misalignment, we apply Singular Vector Canonical Correlation Analysis (SVCCA; \cite{raghu2017svcca}) to the final-layer activation spaces of IID and non-IID models. This analysis aims to determine if the class-discriminative information in non-IID backbones is preserved in a coordinate system that is simply misaligned with the global classification head.

Methodology here is simple. We utilize Singular Vector Canonical Correlation Analysis (SVCCA) to identify a shared linear basis between the IID and non-IID activation spaces. The procedure first applies SVD to the centered activations to whiten the subspaces and remove redundant dimensions, then computes the cross-covariance in this whitened space to find canonical directions that maximize correlation. To test for representational recoverability, we map non-IID test activations into this shared CCA space and reconstruct them back into the IID coordinate system using only the top-$k$ most correlated components. These aligned features are then evaluated using the original, frozen IID classification head.
Finally, these translated' features are passed directly into the original, frozen IID classification head to evaluate their recoverability.

\begin{table}[htbp]
  \caption{Accuracy recovery via SVCCA feature alignment on CIFAR-10. Baseline measures the accuracy of raw non-IID features passed to the original IID head. SVCCA-Aligned measures accuracy after mapping features into the IID coordinate system via the top-$k$ components.}
  \label{tab:svcca_results}
  \centering
  \begin{small}
  \begin{sc}
  \begin{tabular}{lccc}
    \toprule
    Backbone & Baseline (Raw) & SVCCA-Aligned & Components ($k$) \\
    \midrule
    CIFAR-10 CNN    & 12.72\% & 66.26\% & 2048 \\
    CIFAR-10 ResNet & 11.63\% & 50.30\% & 128  \\
    \bottomrule
  \end{tabular}
  \end{sc}
  \end{small}
\end{table}

As shown in Table~\ref{tab:svcca_results}, the original IID classification head performs at near-random levels (~12\%) when fed raw non-IID features. However, after applying a linear SVCCA mapping, accuracy recovers to 66.26\% for the CNN and 50.30\% for the ResNet.

The recovery of accuracy using the frozen IID head confirms that the non-IID backbone retains class-relevant features. The initial failure of the head is attributable to a linear shift in the feature basis—a coordinate misalignment—that renders the stable local representations unreadable to the default classification pathway.

\section{Generalization to Dense Models and Justification for Sparse Training}
\label{app:dense_model_justification}

A potential concern is whether our findings, derived from weight-sparse models, generalize to standard, dense FedAvg. We use a sparse-training regime deliberately as a tool for mechanistic interpretability. Weight-sparse models expose more compact and causally validated circuits, making analysis more tractable and reliable \citep{gao2025weightsparse}. However, to confirm that our conclusions are not artifacts of sparsity, we conducted a parallel set of experiments using a \textbf{dense-model control}. In this control, clients train standard dense models (local magnitude pruning is disabled), which are then averaged by the server. All post-hoc analyses, including circuit discovery, linear probing, and USAE, were performed identically to the sparse-model experiments.

The federated learning setup for the dense control was identical to the sparse experiments, using 10 clients, 100 communication rounds, and 5 local epochs per round for both IID and extreme non-IID ($\alpha=0.05$) partitions. We made two minor adjustments to account for the different optimization landscape of dense models. First, we used a slightly smaller learning rate of $10^{-4}$ (compared to $10^{-3}$ for sparse models) to ensure stable convergence. Second, post-hoc circuit discovery was run for 500 optimization steps (compared to 300 for sparse models) to ensure the gates fully converged on the more complex dense weight structure. All other post-hoc analyses, including linear probing and USAE, were performed identically to the sparse-model experiments.

Our results show that the core mechanistic phenomena persist in the dense setting, confirming that sparsity serves as an effective analytical lens rather than a necessary condition for the observed behaviors.

\begin{figure}[h!]
  \centering
  \begin{subfigure}{0.32\linewidth}
    \centering
    \includegraphics[width=\linewidth]{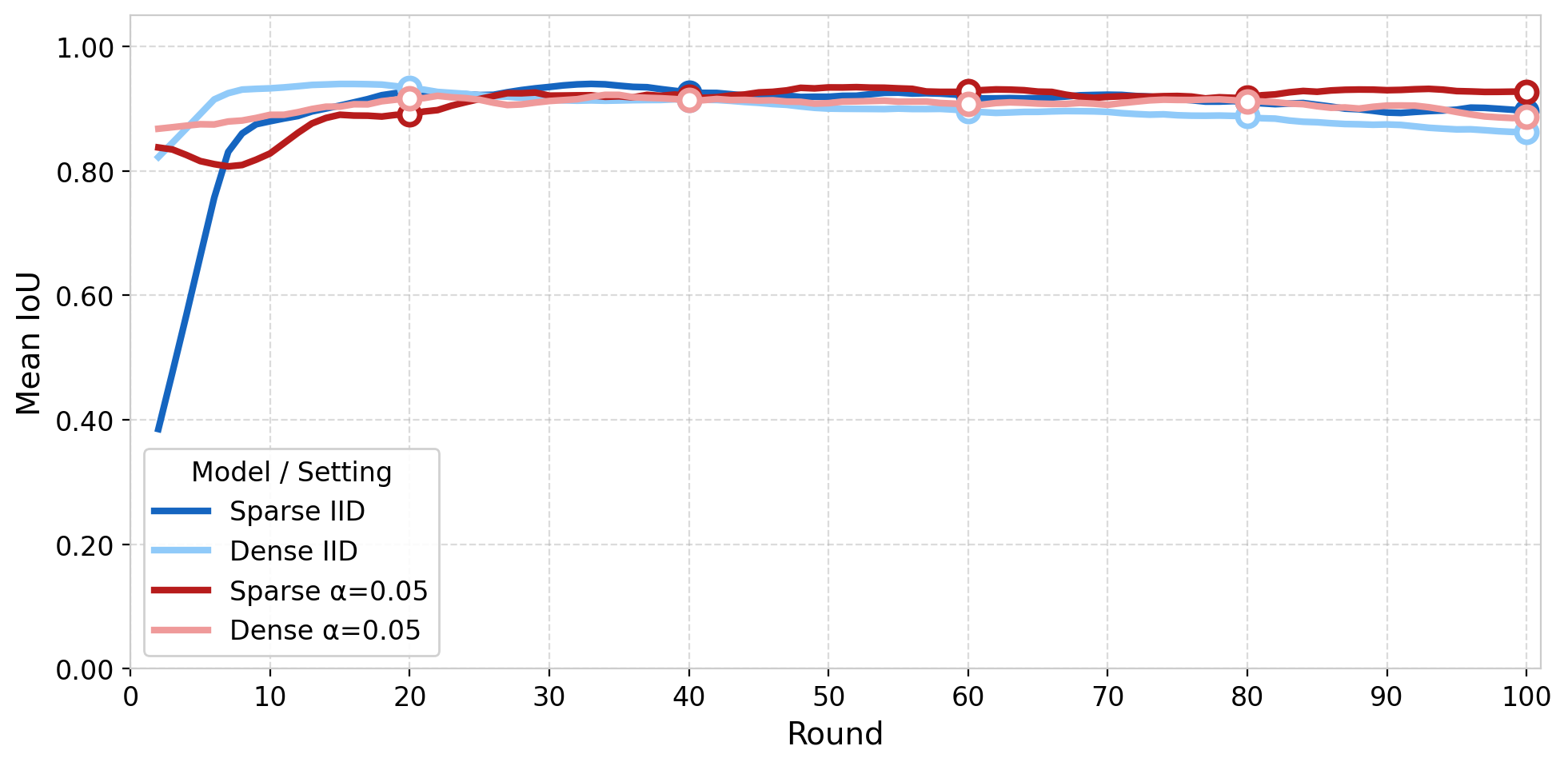}
    \caption{Intra-client Stability (Dense)}
    \label{fig:dense_intra}
  \end{subfigure}
  \hfill
  \begin{subfigure}{0.32\linewidth}
    \centering
    \includegraphics[width=\linewidth]{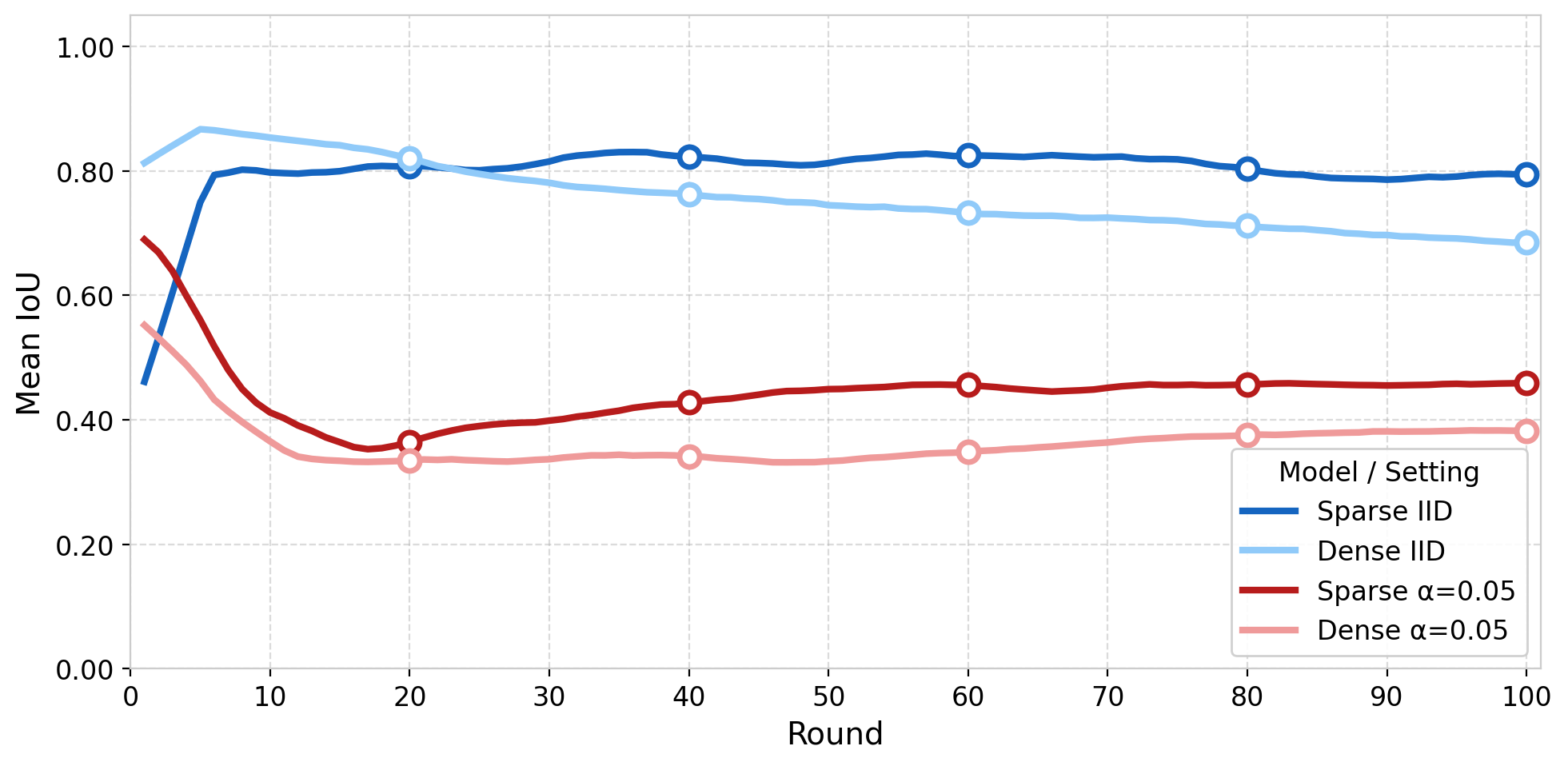}
    \caption{Inter-client Consistency (Dense)}
    \label{fig:dense_inter}
  \end{subfigure}
  \hfill
  \begin{subfigure}{0.32\linewidth}
    \centering
    \includegraphics[width=\linewidth]{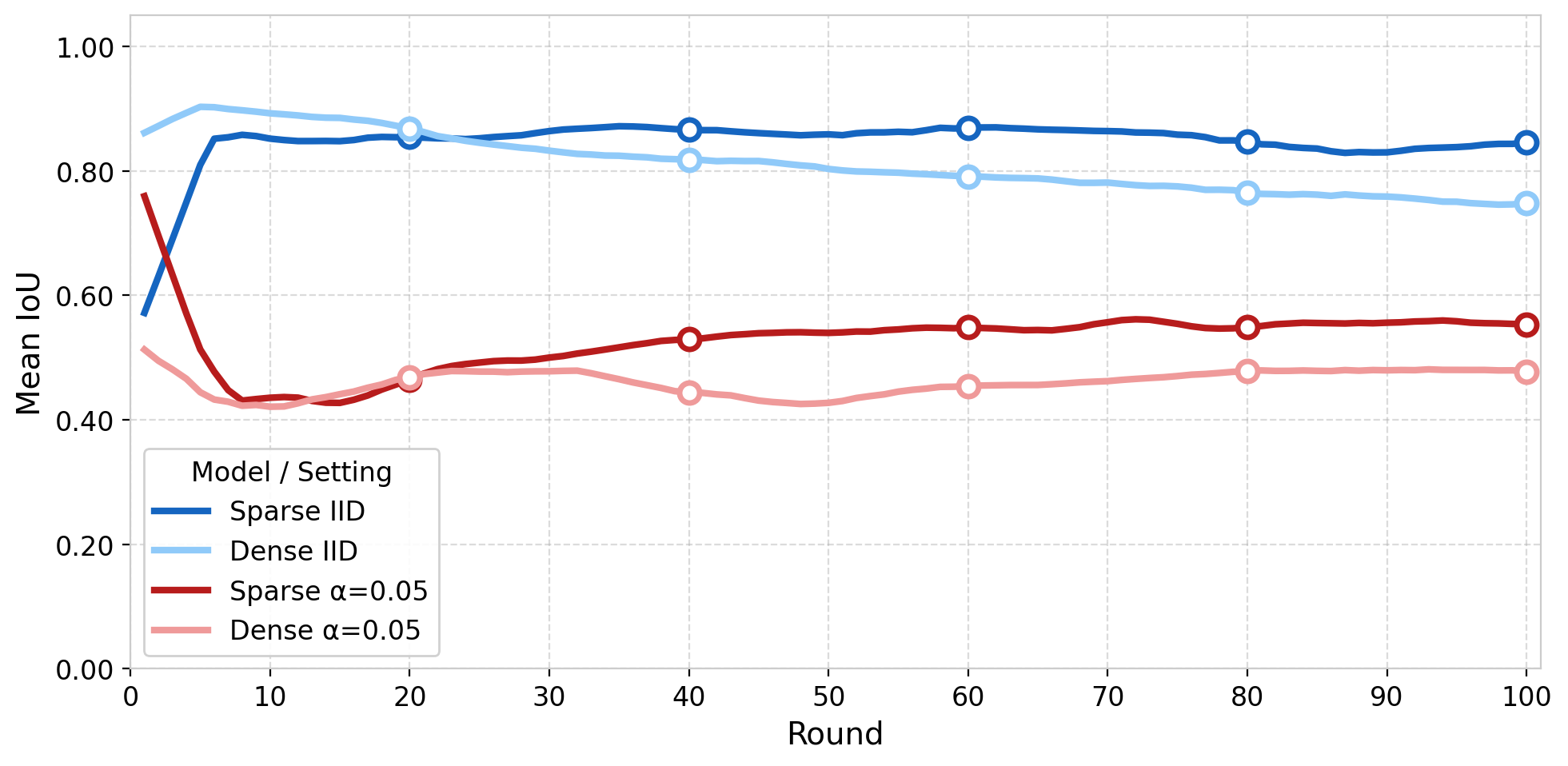}
    \caption{Local vs.\ Global Consistency (Dense)}
    \label{fig:dense_local_global}
  \end{subfigure}
  \caption{Circuit similarity trends for dense models on CIFAR-10 with CNN backbone. The qualitative patterns are identical to the sparse-model results in the main paper. (a) Intra-client stability remains high for both IID and non-IID clients. (b) Inter-client consistency drops sharply under non-IID data. (c) The aggregated global model partially preserves local circuits.}
  \label{fig:dense_circuit_trends}
\end{figure}

\paragraph{Circuit Stability and Divergence.}
As shown in Figure~\ref{fig:dense_circuit_trends}, circuit dynamics in dense models mirror those in sparse models. Intra-client stability remains high (IoU $\approx 0.90$) for both IID and non-IID clients, confirming that local models converge to stable representations. Concurrently, inter-client consistency is high for IID clients (IoU $\approx 0.70$) but collapses for non-IID clients (IoU $\approx 0.38$). The core dynamic of stable but divergent circuits is not dependent on weight sparsity.

\paragraph{Representational Recoverability.}
Table~\ref{tab:dense_diagnostics} confirms that the misalignment between the backbone and classification head persists in dense models. Under extreme Non-IID ($\alpha=0.05$), the dense model's linear probe recovers $+15.9\%$ over global accuracy and head-only finetuning recovers $+10.8\%$, closely matching the sparse model's respective $+21.7\%$ and $+10.2\%$ gains. Both regimes show the same qualitative pattern: features remain substantially more capable than the global head suggests.

\begin{table}[htbp]
  \caption{Representational diagnostics for CIFAR-10 with CNN backbone, comparing sparse and dense models. Linear probe and Head-FT accuracy are measured on frozen backbone features. Delta rows report the gap between Non-IID probe/Head-FT accuracy and global accuracy.}
  \label{tab:dense_diagnostics}
  \centering
  \begin{small}
  \begin{sc}
  \begin{tabular}{llccc}
    \toprule
    Config & Model & Global Acc & Probe Acc & Head-FT Acc \\
    \midrule
    \multirow{2}{*}{IID}
      & Sparse & 68.3\% & 68.5\% & 65.9\% \\
      & Dense  & 74.2\% & 74.2\% & 62.7\% \\
    \midrule
    \multirow{2}{*}{Non-IID ($\alpha=0.05$)}
      & Sparse & 43.1\% & 64.7\% & 53.3\% \\
      & Dense  & 49.9\% & 65.8\% & 60.7\% \\
    \midrule
    \multirow{2}{*}{\textbf{Non-IID Delta}}
      & Sparse & {--} & \textbf{+21.7\%} & \textbf{+10.2\%} \\
      & Dense  & {--} & \textbf{+15.9\%} & \textbf{+10.8\%} \\
    \bottomrule
  \end{tabular}
  \end{sc}
  \end{small}
\end{table}

\paragraph{Shared Feature Basis.}
USAE analysis (Table~\ref{tab:dense_usae}) confirms a high degree of feature sharing in both model types. In both sparse and dense settings, Non-IID features transfer well to the IID head (sparse: 0.5899; dense: 0.6760), closely approaching each model's own same-model upper bound. The Non-IID head fails in both directions regardless of model type, confirming that the classification head is the primary bottleneck rather than the feature representations themselves.

\begin{table}[htbp]
  \caption{USAE concept-transfer accuracy for CIFAR-10 with CNN backbone, comparing sparse and dense models. Same-model entries are reconstruction upper bounds; cross-model entries test concept transfer.}
  \label{tab:dense_usae}
  \centering
  \begin{small}
  \begin{sc}
  \begin{tabular}{llcc}
    \toprule
    Target Head & Source $\to$ Target & Sparse & Dense \\
    \midrule
    \multirow{2}{*}{IID Target}
        & IID $\to$ IID         & 0.6421 & 0.7020 \\
        & Non-IID $\to$ IID     & 0.5899 & 0.6760 \\
    \midrule
    \multirow{2}{*}{Non-IID Target}
        & Non-IID $\to$ Non-IID & 0.4054 & 0.4770 \\
        & IID $\to$ Non-IID     & 0.3645 & 0.4630 \\
    \bottomrule
  \end{tabular}
  \end{sc}
  \end{small}
\end{table}

\end{document}